%% file: main.tex
\documentclass[10pt,twocolumn,letterpaper]{article}

\usepackage{cvpr}              

\input{preamble}

\definecolor{cvprblue}{rgb}{0.21,0.49,0.74}
\usepackage[pagebackref,breaklinks,colorlinks,allcolors=cvprblue]{hyperref}

\usepackage{pifont}
\newcommand{\cmark}{\ding{51}}%
\usepackage[capitalize]{cleveref}
\crefname{section}{Sec.}{Secs.}
\Crefname{section}{Section}{Sections}
\Crefname{table}{Table}{Tables}
\crefname{table}{Tab.}{Tabs.}

\usepackage{pgfplots}
\pgfplotsset{compat=1.18}

\title{QueryOcc: Query-based Self-Supervision for 3D Semantic Occupancy}

\author{
    Adam Lilja$^{1,2}$ \quad
    Ji Lan$^{1,2}$ \quad
    Junsheng Fu$^{2}$ \quad 
    Lars Hammarstrand$^{1}$ \vspace{2mm}\\
    \normalsize$^{1}$Chalmers University of Technology \quad $^{2}$Zenseact\\
    {\tt\small \{firstname.lastname\}@\{zenseact.com, chalmers.se\}}
}

\begin{document}
\twocolumn[{
\maketitle
\begin{center}
    \captionsetup{type=figure}
    \includegraphics[width=\textwidth, trim={0pt 0pt 0pt 0pt}, clip]{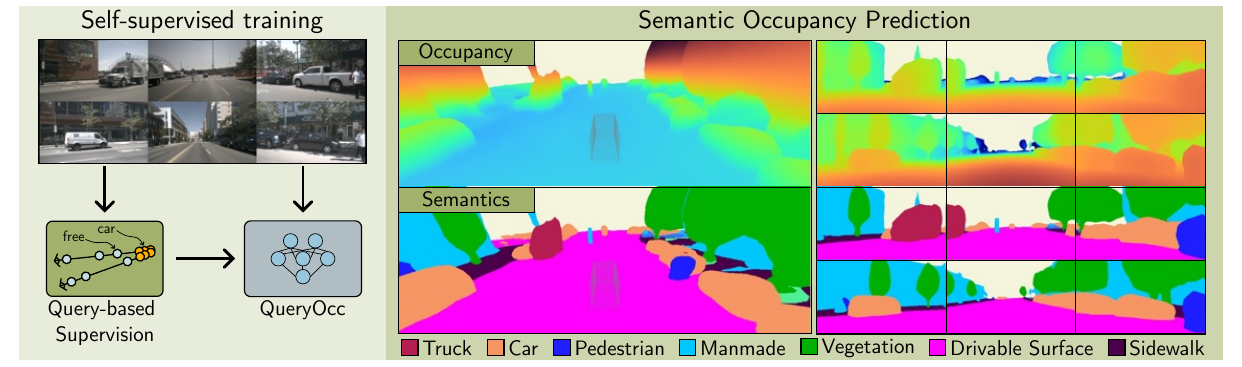}
    \captionof{figure}{\modelname learns to produce continuous 3D semantic occupancy from images through direct spatio-temporal query supervision from sequential frames.
    We outperform prior methods by 26\% in semantic RayIoU while maintaining real-time inference at 11.6 FPS.}
    \vspace{16pt}
    \label{fig:first-fig}
\end{center}
}]
\input{sec/0_abstract}    
\input{sec/1_intro}

\input{sec/2_related_work}
\input{sec/3_method}

\input{sec/4_experiments}

\input{sec/5_conclusions}
\input{sec/6_acknowledgements}
{
    \small
    \bibliographystyle{ieeenat_fullname}
    \bibliography{main}
}

\input{sec/X_suppl}

\end{document}

%% file: preamble.tex
\usepackage{microtype}

\usepackage{xspace}
\newcommand{\modelname}{QueryOcc\xspace}
\newcommand{\modelnameplus}{QueryOcc+\xspace}
\newcommand{\lidar}{lidar\xspace}
\newcommand{\Lidar}{Lidar\xspace}
\newcommand{\supmat}{Sup. Mat.\xspace}
\newcommand{\lidarpcd}{\lidar point cloud\xspace}
\newcommand{\pseudopcd}{pseudo point cloud\xspace}

\newcommand{\anypcd}{point cloud\xspace}

\def\semantic{{Sem.}}
\def\dynamic{{Dyn.}}
\def\occupancy{{Occ.}}

\definecolor{tabfirst}{rgb}{1, 0.7, 0.7} 
\definecolor{tabsecond}{rgb}{1, 0.85, 0.7} 
\definecolor{tabthird}{rgb}{1, 1, 0.7} 

\newcommand{\parsection}[1]{\noindent\textbf{#1:}}
\usepackage{multirow}
\usepackage{textcomp}
\usepackage{tabulary}
\usepackage{colortbl}
\usepackage{siunitx}
\usepackage{graphicx}
\usepackage{nicematrix}

\definecolor{tabfirst}{rgb}{1, 0.7, 0.7} 
\definecolor{tabsecond}{rgb}{1, 0.85, 0.7} 
\definecolor{tabthird}{rgb}{1, 1, 0.7} 
\def\bestrow{\rowcolor{gray!20}}

\usepackage{tikz}
\usetikzlibrary{positioning, fit, calc}
\usepackage{booktabs, xcolor, array}
\newcolumntype{C}[1]{>{\centering\arraybackslash}p{#1}}
\newcolumntype{L}[1]{>{\raggedright\arraybackslash}p{#1}} 

\usepackage{gensymb}
\usepackage{mathtools}

\usepackage[symbol]{footmisc}

\definecolor{CornflowerBlue}{RGB}{100, 149, 237}
\definecolor{YellowGreen}{RGB}{154, 205, 50}
\definecolor{Apricot}{RGB}{251, 206, 177}

\usepackage{comment}

\definecolor{barrier}{RGB}{50, 120, 255}
\definecolor{bicycle}{RGB}{245, 230, 100}
\definecolor{bus}{RGB}{255, 0, 0}
\definecolor{car}{RGB}{245, 150, 100}
\definecolor{construct}{RGB}{255, 0, 0}
\definecolor{motor}{RGB}{150, 60, 30}
\definecolor{pedestrian}{RGB}{30, 30, 255}
\definecolor{traffic}{RGB}{0, 0, 255}
\definecolor{trailer}{RGB}{255, 0, 0}
\definecolor{truck}{RGB}{180, 30, 80}
\definecolor{driveable}{RGB}{255, 0, 255}
\definecolor{other}{RGB}{75, 0, 175}
\definecolor{sidewalk}{RGB}{75, 0, 75}
\definecolor{terrain}{RGB}{80, 240, 150}
\definecolor{manmade}{RGB}{0, 200, 255}
\definecolor{vegetation}{RGB}{0, 175, 0}
\definecolor{others}{RGB}{0, 0, 0}

%% file: sec/0_abstract.tex
\begin{abstract}
Learning 3D scene geometry and semantics from images is a core challenge in computer vision and a key capability for autonomous driving.
Since large-scale 3D annotation is prohibitively expensive, recent work explores self-supervised learning directly from sensor data without manual labels.
Existing approaches either rely on 2D rendering consistency, where 3D structure emerges only implicitly, or on discretized voxel grids from accumulated lidar point clouds, limiting spatial precision and scalability.
We introduce \modelname, a query-based self-supervised framework that learns continuous 3D semantic occupancy directly through independent 4D spatio-temporal queries sampled across adjacent frames.
The framework supports supervision from either pseudo-point clouds derived from vision foundation models or raw lidar data.
To enable long-range supervision and reasoning under constant memory, we introduce a contractive scene representation that preserves near-field detail while smoothly compressing distant regions.
\modelname surpasses previous camera-based methods by 26\% in semantic RayIoU on the self-supervised Occ3D-nuScenes benchmark while running at 11.6 FPS, demonstrating that direct 4D query supervision enables strong self-supervised occupancy learning.
\href{https://research.zenseact.com/publications/queryocc/}{Project page}
\end{abstract}

%% file: sec/1_intro.tex
\section{Introduction}
\label{sec:intro}
Understanding 3D scenes from images is a fundamental challenge in computer vision and a key requirement for autonomous vehicles (AVs).
An AV must reason about geometry, semantics, and free space to plan safely.
To this end, predicting 3D semantic occupancy, a volumetric representation capturing both geometry and semantics, has become an important component of modern AV perception systems \cite{shi2024grid}. 
However, annotating 3D scenes at scale is prohibitively expensive, often requiring resource-intensive labeling of point clouds or multi-view segmentation, and limits the generalization capabilities of these systems.
For instance, annotating semantic occupancy of $850$ sequences in the nuScenes dataset \cite{caesar2020nuscenes} required approximately $4,000$ hours of manual effort \cite{Wang_2023_ICCV}.
This motivates the development of self-supervised approaches that learn directly from sensor data without manual labels.

Existing self-supervised methods fall into two main categories, each with its own limitations:
(1) Rendering-based camera-only approaches \cite{huang2024selfocc, boeder2025gaussianflowocc, wang2024distillnerf} supervise via 2D image reconstruction, enforcing photometric or semantic consistency between views.
While scalable, such training provides only indirect geometric signals and 3D structure emerges as a by-product of image synthesis rather than as an explicit learning target.
Moreover, they often rely on externally estimated or pseudo depth maps for stable training and strong performance~\cite{jiang2025gausstr, boeder2025gaussianflowocc}.
(2) Lidar-based approaches \cite{sze2025minkocc, vobecky2023pop} provide explicit and direct supervision in 3D, but rely on discretizing aggregated lidar point clouds into voxel grids of predefined range and resolution for learning, which limits spatial precision and scalability.

We hypothesize that direct supervision in continuous 4D space-time provides clearer geometric feedback for self-supervised 3D semantic occupancy learning than rendering-based or voxelized lidar methods.
To this end, we introduce \emph{\modelname}, a flexible framework that can leverage geometric supervision from either real lidar or point clouds derived from camera images using a depth prediction model.
Together with semantic pseudo-labels from a vision foundation model, these point clouds provide joint supervision of geometry and semantics through independent 4D queries sampled across adjacent frames. 

We also argue that self-supervised learning should leverage long-range supervision and remain faithful to the unbounded real world.
To achieve this, we propose a compact and contractive scene representation that preserves near-field detail while smoothly compressing distant regions.
In addition, we design a mechanism, \emph{lift-contract-splat}, that lifts image features into this contracted space. 
These components together allow supervision over the full observable scene under constant memory and computation, while supporting both camera-only and camera-lidar supervision.

\noindent In summary, our main contributions are:
\begin{itemize}
\item introducing a flexible query-based framework \textbf{\modelname}, that learns continuous 3D semantic occupancy from multi-view images without rendering losses or lidar aggregation and voxelization,
\item designing an \textbf{unbounded contractive scene representation} that allows for long-range supervision, preserves near-field detail while compressing far-field, and proposing a mechanism to lift image features into this representation, 
\item demonstrating \textbf{state-of-the-art performance} on the self-supervised Occ3D-nuScenes benchmark, outperforming prior methods by 26\% in semantic RayIoU, while maintaining real-time inference at 11.6 FPS, and highlighting the efficacy of direct 4D query-based self-supervision.
\end{itemize}

%% file: sec/2_related_work.tex
\begin{figure*}[!t]
    \centering
    \includegraphics[width=\textwidth, trim={0pt 0pt 0pt 0pt}, clip]{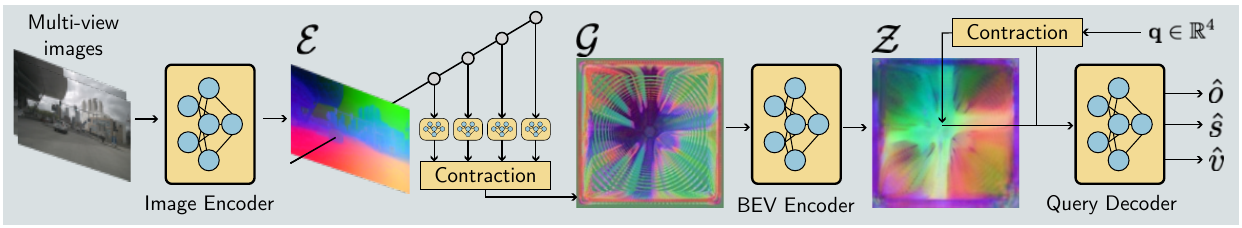}
    \caption{
    Overview of \modelname. 
    Multi-view camera images are encoded and lifted to BEV via our lift-contract-splat module, combining geometric encoding, log-linear depth bins, and an axis-aligned BEV contraction. 
    The BEV features form a spatially grounded representation from which a unified decoder predicts occupancy, semantics, or distilled vision foundation model features for continuous queries.
    }
    \label{fig:method}
\end{figure*}

\section{Related Work}
\label{sec:related-work}
Research on 3D occupancy prediction has evolved from fully supervised voxel-based methods to self-supervised frameworks aiming to learn directly from sensor data.
Our work builds upon camera-based occupancy learning, 3D scene representations, and query-based learning, bridging these in an unified self-supervised formulation.

\subsection{3D Semantic Occupancy Prediction}
MonoScene \cite{cao2022monoscene} pioneers voxel-wise occupancy prediction from monocular images, and OccDepth \cite{miao2023occdepth} extends this approach through implicit depth information from stereo images. 
Subsequent works \cite{huang2023tri, wei2023surroundocc} adopt multi-camera inputs, explore modality fusion \cite{Duan_2025_CVPR}, efficient scene representations \cite{huang2024gaussianformer, huang2025gaussianformer}, and foundation-model priors \cite{xue2025sdformer}. 
However, while all the above methods supervise directly in 3D, they rely on voxel-based 3D ground-truth supervision \cite{tian2023occ3d,Wang_2023_ICCV}, which is expensive to obtain and limits scalability. 
To overcome this limitation, recent work explores self-supervised learning, differing mainly in the type of supervision signal (2D- or 3D-based) and the underlying scene representation.

\parsection{2D-based self-supervision}
Rendering-based methods such as SelfOcc~\cite{huang2024selfocc}, OccNeRF~\cite{zhang2023occnerf}, and LangOcc~\cite{boeder2025langocc} treat their latent volumes as radiance fields and adopt NeRF-style volume rendering~\cite{mildenhall2021nerf} to reconstruct images of the scenes.
Scene geometry is learned implicitly through photometric consistency, while semantics come from 2D pseudo-labels generated by vision foundation models such as OpenSeeD~\cite{zhang2023simple}, GroundedSAM~\cite{ren2024grounded}, or MaskCLIP~\cite{zhou2022extract}.
Subsequent works replace neural volume rendering with Gaussian splatting for efficiency~\cite{gan2025gaussianocc, jiang2025gausstr}.
GaussianFlowOcc~\cite{boeder2025gaussianflowocc} further improves scalability through attention optimization and temporal supervision.
While these methods scale well, they provide only indirect geometric supervision through 2D supervision.
We supervise directly in 3D, allowing geometry and semantics to be learned as explicit targets rather than by-products of rendering.

\parsection{3D-based self-supervision}
A complementary line of research uses \lidar signals to obtain explicit 3D supervision.
MinkOcc~\cite{sze2025minkocc} fuses camera and \lidar inputs and supervises geometry using voxelized \lidar accumulations.
Similarly, POP3D~\cite{vobecky2023pop} projects \lidar points into images to associate them with CLIP features, constructing a voxelized 3D feature grid that the model learns to reproduce.
\modelname generalizes this direction by learning directly from continuous points without aggregation or voxelization.

\parsection{3D scene representation}
A central challenge in 3D perception is how to lift 2D image features into spatial representations that balance geometric fidelity and computational efficiency.
Existing self-supervised methods mainly differ in this choice of representation, which determines how geometry is encoded and supervised.
\textit{\textbf{Grid-based}} methods project multi-view features into discretized 3D voxel grids, where fusion and learning occur directly in 3D~\cite{zhang2023occnerf, boeder2025langocc, gan2025gaussianocc, vobecky2023pop}.
This provides explicit volumetric reasoning but scales poorly with spatial range and requires heavy 3D convolutions.
Bird’s-Eye View (BEV)~\cite{philion2020lift} collapses the vertical axis into a single ground-plane projection, achieving high efficiency and Tri-Perspective View (TPV)~\cite{huang2023tri} collapses the three spatial axes separately into orthogonal planes, recovering some 3D structure but at the cost of triple feature processing and limited spatial consistency across planes.
Grid-based methods also differ in how 2D features are lifted into 3D space.
Pull-based approaches~\cite{li2024bevformer} project 3D points into the image plane to sample features, while push-based methods~\cite{philion2020lift} estimate per-pixel depth distributions and project features into 3D.
The former tend to limit spatial separation, and the latter are generally bounded by predefined depth ranges, limiting coverage of unbounded scenes.
\textit{\textbf{Sparse point primitives}} such as 3D Gaussians~\cite{jiang2025gausstr, boeder2025gaussianflowocc} replace dense voxel volumes with point-based representations to improve efficiency, but depend on explicit point storage, attention across many points, and 2D rendering-based supervision, leading to indirect geometric learning.
\modelname retains BEV's computational efficiency and compact representation, extends it to unbounded scenes through spatial contraction, and recovers full continuous 3D expressiveness by using a lightweight query-based decoder.

\subsection{Query-based Learning}
Query-based formulations predict scene properties such as occupancy at arbitrary points instead of dense voxel grids, enabling continuous implicit representations.
ImplicitO~\cite{agro2023implicit} applies this to \lidar perception by modeling occupancy and flow as a continuous function.
Other works adopt the formulation for self-supervised pre-training for \lidar perception:
ALSO~\cite{boulch2023also} reconstructs continuous occupancy from sampled \lidar points,
UnO~\cite{agro2024uno} extends to forecasting future occupancy,
and GASP~\cite{ljungbergh2025gasp} unifies geometric, feature distillation, and ego-path supervision.
Together, these methods exemplify the query-based paradigm for \lidar-driven perception.
In contrast, our work brings query-based learning to camera-only semantic occupancy, addressing the challenges of noisy image-derived geometry, semantic supervision, and efficient long-range reasoning via a contractive BEV representation.

%% file: sec/3_method.tex
\section{Method}
\label{sec:method}
We propose \modelname, a self-supervised framework for learning continuous 3D semantic occupancy from multi-view images through spatio-temporal query-based supervision.
Given $m$ multi-view images with known camera calibrations $\mathbf{M}_t = \{ I_i \}_{i=1}^m$ at the current timestep and any 3D point at time $t$, here referred to as \emph{query}, $\mathbf{q} = [x, y, z, t]^\top$, 
the model $\mathrm{F}_{\theta}$ predicts whether the query is occupied $o \in \{0, 1\}$ and its semantic label $s \in \{1, \ldots, N_s\}$, where $N_s$ is the number of classes.
Formally, the model predicts the probability distribution:
$\hat{o}(\mathbf{q}) = Pr\{o(\mathbf{q})=1 \mid \mathbf{M}_t)$, 
and conditional distribution over the classes, 
$\mathbf{\hat{s}}(\mathbf{q}) = Pr\{s(\mathbf{q})  \mid o(\mathbf{q})=1, \mathbf{M}_t)$, 
such that: 
$\langle \hat{o}, \mathbf{\hat{s}} \rangle = F_{\theta}(\mathbf{M}_t, \mathbf{q})$.
To generate supervision, we use adjacent frames
$\mathbf{M}_T = \{\mathbf{M}_\text{Tmin}, \ldots, \mathbf{M}_\text{Tmax}\}$, 
with known ego poses.  
We next describe the model architecture and unbounded scene representation, the direct query-based supervision scheme, and training details.

\subsection{Model Overview}
\label{sec:method:model}
\modelname follows a modular design consisting of four stages: 
(1) image encoding $\mathcal{E} = \mathrm{E}_\psi(\mathbf{M}_t)$, 
(2) lifting image features to contractive BEV, $\mathcal{G} = \mathrm{G}_\xi(\mathcal{E})$, 
(3) processing the BEV features $\mathcal{Z} = \mathrm{H}_\zeta(\mathcal{G})$, and 
(4) query-based decoding, where occupancy and semantics are predicted for arbitrary queries $\mathbf{q}$ through a unified decoder, $\langle \hat{o}, \mathbf{\hat{s}}\rangle = \mathrm{D}_\phi(\mathcal{Z}, \mathbf{q})$.
The entire pipeline is illustrated in \cref{fig:method}, and can be expressed compactly as:
\begin{equation}
    \langle \hat{o}, \mathbf{\hat{s}} \rangle = \mathrm{F}_{\theta}(\mathbf{M}_t, \mathbf{q}) = \mathrm{D}_\phi\biggl(\mathrm{H}_\zeta\Bigl(\mathrm{G}_\xi\bigl(\mathrm{E}_\psi(\mathbf{M}_t)\bigl)\Bigl), \mathbf{q}\biggl)
\end{equation}
where $\theta = \{\psi, \xi, \zeta, \phi\}$ denotes the set of all learnable parameters.
The model is trained end-to-end using the proposed self-supervised scheme described in \cref{sec:method:supervision}.

\parsection{Image encoder $\mathrm{E}_\psi$}
Each input image is processed independently using a pre-trained standard backbone \cite{resnet, convnext} to extract dense per-view features.
Camera calibration parameters (intrinsics and extrinsics) are encoded using a lightweight MLP and are applied to the image features, enabling the model to learn camera-specific details.

\parsection{Lift-contract-splat $\mathrm{S}_\xi(\mathcal{E})$}
A central goal of \modelname is to encode unbounded 3D scenes efficiently while maintaining geometric fidelity near the ego vehicle.
To achieve this, we design a \emph{Lift–Contract–Splat} module that extends the Lift-Splat-Shoot (LSS) formulation \cite{philion2020lift} with three key innovations:
(1) a point-wise encoding for explicit geometric reasoning,
(2) adaptive log-linear depth binning for balanced near–far resolution, and
(3) an axis-aligned spatial contraction for unbounded scene coverage.

\textbf{\textit{Base formulation}}
Following LSS, each pixel feature $\boldsymbol\epsilon$ in $\mathcal{E}$ is associated with a categorical depth probability distribution $p_d$ over $N_d$ bins.
Given known intrinsics and extrinsics, each pixel is lifted into $N_d$ 3D points in the ego frame, each weighted by its depth-bin probability.
Point clouds from all camera views are then fused by splatting the resulting 3D feature cloud onto a shared BEV grid.

\textbf{\textit{(1) Point encoding:}}
Instead of splatting raw features $f_d = p_d \cdot \boldsymbol\epsilon$, we explicitly encode visibility, uncertainty, and temporal cues through a learnable function $P_\pi$:
$f_d = P_\pi\big(p_d \cdot \boldsymbol\epsilon + (1 - p_d) \mathbf{e},\, p_v,\, p_d, t_c \big),$
where $\mathbf{e}$ is a learnable empty-space embedding and $p_v$ is the cumulative depth probability up to the depth bin $d$ (interpreted as the predicted visibility) and $t_c$ is the timestamp of the current image.
$P_\pi$ applies Fourier feature encoding to each input and aggregates them through a lightweight MLP.
This design enables the model to represent both occupied and unobserved regions explicitly, improving geometric reasoning in the lifted feature space.

\textbf{\textit{(2) Log-linear depth bins:}}
To emphasize high-resolution geometry near the vehicle while enabling coverage at a distance without increasing the number of depth bins, we replace the uniform binning with exponentially increasing spacing outside the $d_{\text{near}}$ region:
\begin{equation}
\label{eq:exponential-depth-bins}
d(r) = (1 - \alpha)\, d_{\text{near}} \!\left(\frac{d_{\text{far}}}{d_{\text{near}}}\right)^{r} \!\! + 
\alpha\, [d_{\text{near}} + r (d_{\text{far}} - d_{\text{near}})],
\end{equation}
where $r \in [0, 1]$ is the depth normalized with the maximum range $d_{\text{far}}$, and $\alpha$ balances near- vs far-field coverage.

\textbf{\textit{(3) Contraction for unbounded scenes:}}
To maintain constant memory as range increases, we introduce an axis-aligned BEV contraction inspired by NeRF-based scene parameterizations \cite{barron2022mip, wang2024distillnerf, tonderski2024neurad}.
Formally, we define a continuous contraction function $f_{\text{contr.}} : \mathbb{R} \rightarrow [-1, 1]$ around the ego vehicle that preserves coordinate continuity at the high-resolution border $\bar{\kappa}=1$:
\begin{equation}
\label{eq:bev_contraction}
f_{\text{contr.}}(\bar{\kappa}) =
\begin{cases}
\beta \cdot\bar{\kappa}, & |\bar{\kappa}| \le 1,\\
\mathrm{sign}(\bar{\kappa})\left(1 - \frac{1-\beta}{|\bar{\kappa}|}\right), & |\bar{\kappa}| > 1,
\end{cases}.
\end{equation}
where the coordinate is normalized $\bar{\kappa} = \frac{\kappa}{K_{hr}}$ for $\kappa \in \{x, y\}$ with the high resolution range $K_{hr}$, and $\beta$ sets the ratio of the BEV grid used for high resolution.
Unlike spherical contraction \cite{barron2022mip, wang2024distillnerf}, our axis-aligned variant preserves rectangular geometry while smoothly compressing distant coordinates.

\begin{figure}[t]
  \centering
    \includegraphics[width=0.8\linewidth]{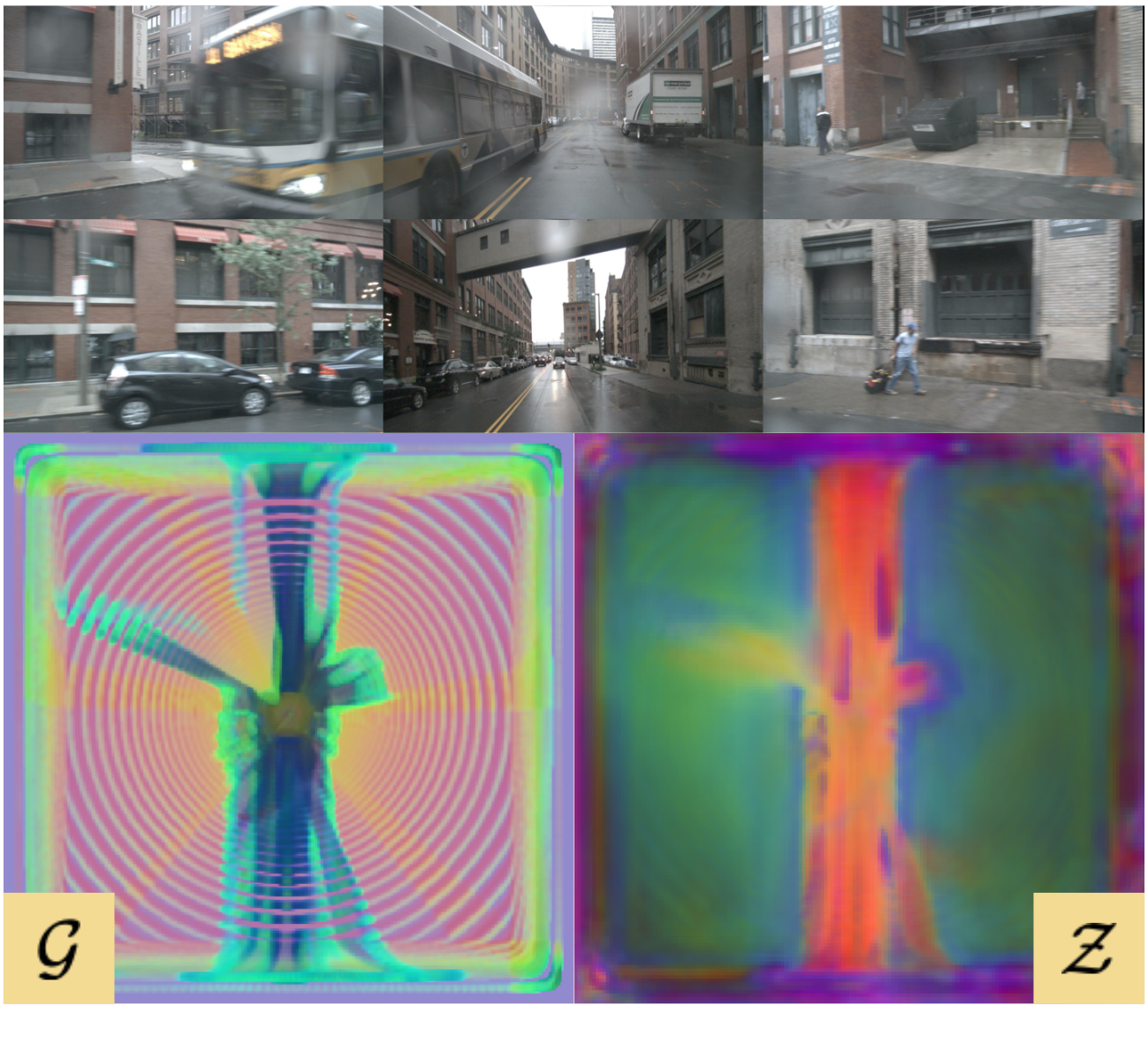}
  \caption{
  PCA visualization of lifted features $\mathcal{G}$ and BEV features $\mathcal{Z}$.
  The proposed BEV contraction and point encoding enable efficient modeling of large scenes, and the learned BEV representations exhibit structured separation between occupied, occluded, and free-space areas. }
  \label{fig:contraction-features}
\end{figure}

The contraction enables the model to encode distant structures within a fixed BEV grid while preserving high spatial resolution near the ego vehicle. 
This is important because BEV size, and thus memory and computation, otherwise grows quadratically with scene range. 
Distant areas contain sparse geometry and uncertain depth estimates, compressing them yields efficiency gains with negligible loss in fidelity. \cref{fig:contraction-features} shows an example of the learned contracted features.

\parsection{BEV feature processing $H_\zeta(\mathcal{S})$}
After splatting the features to BEV, we process the resulting feature map using a series of ResBlocks \cite{agro2024uno} and multi-scale deformable attention \cite{zhu2021deformable} to capture both local and global spatial dependencies.
We apply dynamic convolutions to enable adaptation to the local scaling of the contraction.

\parsection{Unified decoder for spatio-temporal queries}
The contracted BEV feature map $\mathcal{Z}$ represents a spatially grounded field, shown in \cref{fig:contraction-features}, from which we can query the scene at any 4D point, or \emph{query}, $\mathbf{q} = [x, y, z, t]^\top$.
We propose a unified multi-task lightweight decoder that shares parameters across all prediction targets, \eg, occupancy, semantics, and vision foundation model (VFM) feature distillation.

\begin{figure*}
    \centering
    \includegraphics[width=\linewidth, trim={0pt 5pt 0pt 0pt}, clip]{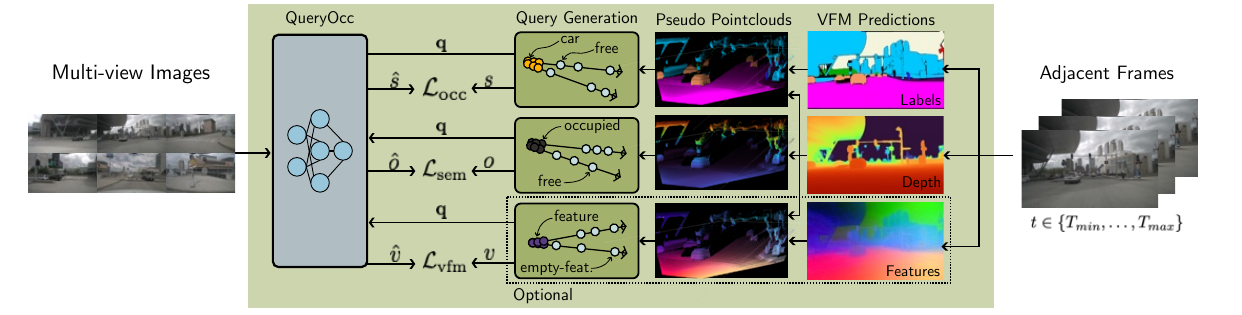}
    \caption{
    Overview of the self-supervised supervision process for a camera-only setup.
    Adjacent frames provide supervision through pseudo point clouds from VFM-predicted depth, semantic pseudo-labels, or features.
    These 3D points generate positive and negative 4D queries used to supervise occupancy, semantics, and feature distillation.
    The framework can optionally be complemented by \lidarpcd.
    }
    \label{fig:supervision}
\end{figure*}

Each query $\mathbf{q}$ is first contracted using \cref{eq:bev_contraction} to align with the contracted BEV space and allow model supervision at longer ranges.
The query is encoded by a small MLP and fused with interpolated BEV features at the corresponding spatial position.
To maintain spatial adaptivity, the decoder follows \cite{agro2024uno, ljungbergh2025gasp} and predicts offsets $\Delta \mathbf{q}_{x,y}$ to additional extraction of features from the BEV feature map at nearby locations.
This offset-based aggregation mimics the dynamic spatial reasoning of deformable attention, allowing the model to focus on locally relevant regions. 
The aggregated features are iteratively refined through shallow residual layers before producing a multi-task output: $\mathbf{\hat{a}} = \mathrm{D}_\phi(\mathbf{q}, \mathcal{Z})$
where $\mathbf{a}$ is determined by the supervision signals used. 
In the standard configuration, the decoder predicts occupancy and semantics, $\mathbf{\hat{a}} = \langle \hat{o}, \mathbf{\hat{s}} \rangle$.
The formulation can be extended to include feature distillation by adding distillation of a VFM feature vector $\mathbf{\hat{v}}$, yielding target $\mathbf{\hat{a}} = \langle \hat{o}, \mathbf{\hat{s}}, \mathbf{\hat{v}} \rangle$, see \cref{sec:method:supervision}.

This shared decoding formulation unifies geometric, semantic, and high-level visual supervision within a single differentiable query framework, promoting both computational efficiency and cross-task consistency.

\subsection{Self-Supervised Training}
\label{sec:method:supervision}
Self-supervised learning in \modelname operates directly in continuous 4D space-time, providing supervision without going through rendering losses or relying on pseudo-label pipelines to aggregate and voxelize lidar data.
In our framework, \cref{fig:supervision}, supervision is generated from real or pseudo 3D point clouds observed across time, which we unify into a single 4D query-based formulation described below.

\parsection{Supervisory signal sources}
We first require point clouds with associated semantic labels or vision foundation model (VFM) features.
The point clouds are obtained either from images or from lidar, depending on sensor availability.

\textbf{Camera-only}
When lidar is unavailable, we construct pseudo point clouds $\mathcal{P}_{\text{pseudo}}$ by lifting pixels into 3D using metric depth predicted by an off-the-shelf vision depth model using known camera intrinsics and extrinsics.
Each lifted point is paired with semantic pseudo-labels or dense VFM features extracted from the same images, providing geometric and semantic targets in ego coordinates.
This is the minimal required set-up and we refer to it as \emph{\modelname}.

\textbf{Camera-lidar}
If \lidar data is accessible, the observed point cloud $\mathcal{P}_{\text{lidar}}$ can be used as a source of explicit geometric supervision.
Each lidar point is projected into the corresponding camera views to retrieve its semantic label or feature embedding from the image-based VFM outputs.
We follow~\cite{ljungbergh2025gasp} and apply min-depth filtering to mitigate the worst artifacts from occlusions and calibration errors.

\textbf{Unified}
Finally, both modalities can be combined: $\mathcal{P}_{\text{uni}} = \mathcal{P}_{\text{pseudo}} \cup \mathcal{P}_{\text{lidar}}$. 
This allows any mixture of pseudo- and \lidar-based supervision, enabling geometry and semantics to be learned jointly from the available data.
This is an extended set-up and we refer to it as \emph{\modelnameplus}.

\parsection{4D query generation}
The point clouds serve as empirical evidence from which supervision signals are sampled.
Supervision is applied at the level of spatio-temporal 4D \emph{queries}, $\mathbf{q}$, generated from a \anypcd $\mathcal{P}_{\text{pseudo}}$, $\mathcal{P}_{\text{lidar}}$, or $\mathcal{P}_{\text{uni}}$. 
Each query is associated with an occupancy label $o$, semantic target $\mathbf{s}$, and/or a high-dimensional feature embedding $\mathbf{v}_{\text{vfm}}$ from a VFM.
We thus produce a set of $N$ data samples $\mathcal{D} = \{\langle\mathbf{q}_i, \mathbf{a}_i \rangle\}_{i=0}^N$ consisting of queries $\mathbf{q}_i$ and targets $\mathbf{a}_i$ from adjacent data at $t \in \{T_{min}, \ldots,  T_{max}\}$. 
We assume each point in the \anypcd $\mathbf{p_i} \in \mathcal{P}$ was captured from a sensor with origin $\mathbf{o}_i$, and each point ${\mathbf{p_i}} = (x_i,y_i,z_i)$ and $\mathbf{o}_i = (x_i, y_i, z_i)$ has a corresponding capture time $t_i$.
Negative, \emph{Unoccupied}, queries are sampled along the ray from the sensor origin up to the point:
\begin{equation}
    \label{eq:negative-queries}
    \mathcal{D}^- = \{\langle \mathbf{o}_i + r (\mathbf{p}_i - \mathbf{o}_i) , \mathbf{0}\rangle \ | \  r \in (0, 1)  \}_{i=0}^N
\end{equation}
Positive, \emph{occupied}, queries are generated within a buffer zone with length $\delta$ behind the point:
\begin{equation}
    \label{eq:positive-queries}
     \mathcal{D}^+ =  \{ \langle \mathbf{p}_i + \frac{r(\mathbf{p}_i - \mathbf{o}_i)}{||\mathbf{p}_i - \mathbf{o}_i||} , \mathbf{a}_i \rangle \ | \  r \in (0, \delta) \}_{i=0}^N
\end{equation}
We sample a balanced set of positive and negative queries across space and time.
These independent and continuous queries provide sparse but direct supervision without discretization or post-processing, enabling scalable self-supervised learning of semantic occupancy.

\parsection{Losses}
Each query contributes to one or more losses depending on the available supervision:
\begin{align}
\mathcal{L} =
\lambda_{\text{occ}}\mathcal{L}_{\text{occ}}(\hat{o}(\mathbf{q}), o) +
\lambda_{\text{sem}}\mathcal{L}_{\text{sem}}(\hat{s}(\mathbf{q}), \mathbf{s}) \nonumber \\ + 
\lambda_{\text{vfm}}\mathcal{L}_{\text{vfm}}(\hat{v}(\mathbf{q}), \mathbf{v}),
\end{align}
where $\mathcal{L}_{\text{occ}}$ is a binary cross-entropy loss, $\mathcal{L}_{\text{sem}}$ is a categorical cross-entropy over semantic classes, and $\mathcal{L}_{\text{vfm}}$ is L1.

\subsection{Training Details}
\label{sec:method:training}
For a fair comparison, we align key hyperparameters with the strongest baseline, GaussianFlowOcc \cite{boeder2025gaussianflowocc}.
We therefore adopt an input resolution of $224\times704$ pixels and use Metric3D \cite{yin2023metric3d} to provide supervision for occupancy $o$, and Grounded-SAM \cite{ren2024grounded} for semantic pseudo-labels $s$ for all main experiments.
We refer to this model as \textbf{\modelname}.
A full training run completes in approximately $13$ hours on $4$ A100 GPUs, requiring about $30$ GB of GPU memory. 
To push the boundaries for our framework we also train a model using both camera and lidar data for supervision, distillation of VFM features, and a larger input resolution of $900\times1600$ and refer to this set-up as \textbf{\modelnameplus}.
For VFM supervision of $\mathbf{v}_{\text{vfm}}$ we use DinoV3 \cite{simeoni2025dinov3} ViT-base features reduced to $16$ dimensions using PCA reduction matrix from a subset of the training data.
For both models, we use a ConvNeXt-Base \cite{convnext} as image encoder, but see similar results and state-of-the-art performance also with \eg a ResNet-50 \cite{resnet} in \supmat \cref{sec:supmat:experiment-details}.
We set LCS parameters $\alpha = 0.3$, $d_{\text{far}} = 100$m, and $d_{\text{near}} = 40$m, and BEV contraction $K_{hr}=40$, $\beta=0.8$.
We use $30k$ point cloud points, generating $800$k queries per sample.
See \supmat \cref{sec:supmat:experiment-details} for full experimental setup.

%% file: sec/4_experiments.tex
\begin{table}[t]
\centering
\setlength{\tabcolsep}{2pt} 
\renewcommand{\arraystretch}{0.7} 
\begin{tabular}{
    l
    C{0.7cm} C{0.7cm} C{0.7cm} 
    C{0.7cm} C{0.7cm} C{0.7cm} 
}
\toprule
\multirow{2}{*}{\textbf{Method}} 
 & \multicolumn{3}{c}{\textbf{RayIoU}$\uparrow$} 
 & \multicolumn{3}{c}{\textbf{IoU}$\uparrow$} \\ 
\cmidrule(lr){2-4} \cmidrule(lr){5-7}
 & \semantic & \dynamic & \occupancy 
 & \semantic & \dynamic & \occupancy \\
\midrule
SelfOcc$^{\dagger}$~\cite{huang2024selfocc}    & 10.9 & 7.2  & 29.2 & 10.5 & 3.7  & 45.0 \\
OccNeRF~\cite{zhang2023occnerf}    & --   & --   & --   & 10.8 & 3.7  & 22.8   \\    
DistillNeRF~\cite{wang2024distillnerf}   & --   & --   & --   & 10.1 & 5.2  & 29.1   \\
LangOcc$^{\dagger}$~\cite{boeder2025langocc}             & 11.6 & 9.0  & \cellcolor{tabsecond}38.7 & 13.3 & 7.7  & \cellcolor{tabsecond}51.8 \\
GaussianOcc~\cite{gan2025gaussianocc}          & 11.9 & --   & --   & 11.3 & 7.0   & --   \\
MinkOcc~\cite{sze2025minkocc} {\footnotesize-w/lidar}        & 12.5 & --   & --   & 13.2 & 3.4   & --   \\
GaussTR$^{\dagger}$~\cite{jiang2025gausstr} {\footnotesize-FeatUp}    & 13.8 & 14.5 & 34.2 & 13.3 & 9.0  & 45.2 \\
GaussTR$^{\dagger}$~\cite{jiang2025gausstr} {\footnotesize-T2D}        & \cellcolor{tabthird}14.2 & \cellcolor{tabsecond}17.7 & 33.8 & \cellcolor{tabthird}13.9 & \cellcolor{tabfirst}13.4  & 44.5 \\

GaussianFlowOcc~\cite{boeder2025gaussianflowocc}      & \cellcolor{tabsecond}18.7 & --   & --   & \cellcolor{tabsecond}17.1 & \cellcolor{tabthird}10.1 & \cellcolor{tabthird}46.9 \\
GaussianFlowOcc*     & 18.2 & \cellcolor{tabthird}17.2 & \cellcolor{tabthird}36.0 & 16.1 & 9.9 & 40.2 \\
\textbf{\modelname}  & \cellcolor{tabfirst}23.6 & \cellcolor{tabfirst}21.7 & \cellcolor{tabfirst}45.2
                     & \cellcolor{tabfirst}21.3 & \cellcolor{tabsecond}13.2 & \cellcolor{tabfirst}55.0 \\ \midrule
\textbf{\modelnameplus} &\textbf{25.8} & \textbf{23.8} & \textbf{47.4} & \textbf{23.5} & \textbf{15.7} & \textbf{56.9} \\
\bottomrule
\end{tabular}
\caption{\modelname sets a new state of the art among self-supervised methods on Occ3D-nuScenes. 
\modelnameplus uses high-res images as input, camera and lidar for supervision, and VFM feature distillation supervision, pushing the limits for our framework and performs \textbf{best} of all methods. 
Colors indicate \colorbox{tabfirst!75}{First}, \colorbox{tabsecond!75}{Second}, \colorbox{tabthird!75}{Third} in ranking among more similar methods.
$^{\dagger}$ denotes reproduction for RayIoU, and * denotes reproduction for both RayIoU and IoU. }
\label{tab:main_results}
\end{table}

\section{Experiments}
\label{sec:experiments}
We evaluate \modelname on semantic occupancy prediction, quantify effects of core components, and examine scalability with respect to supervision source, resolution, and data scale.

\subsection{Dataset and Evaluation}
Following standard practice in recent work~\cite{jiang2025gausstr, boeder2025gaussianflowocc, boeder2025langocc, gan2025gaussianocc}, 
we evaluate on the Occ3D-nuScenes benchmark~\cite{tian2023occ3d}, which provides ground-truth semantic occupancy derived from the nuScenes dataset~\cite{caesar2020nuscenes}. 
The environment around the ego vehicle is discretized into voxels of edge length $0.4$\,m, covering $\pm40$\,m laterally and longitudinally and $[-1, 5.5]$\,m vertically.
While \modelname learns a continuous semantic representation, we here query it in the Occ3D-voxel centers.

We report intersection-over-union (IoU) for both binary occupancy (\occupancy) and semantic prediction (\semantic). 
In addition, we use the more robust \emph{RayIoU}, which evaluates per-ray depth consistency and penalizes geometrically inflated surfaces ~\cite{liu2024fully}. 
For a more targeted semantic evaluation, we report both the mean over all classes and mean computed over the eight dynamic classes (denoted \dynamic), \eg pedestrians, cars, trucks, out of the fifteen total semantic categories. 
Class definitions and full class-wise results are provided in \supmat \cref{sec:supmat:dataset} and \cref{sec:supmat:results} respectively.

\begin{table}[]
    \centering
    \setlength{\tabcolsep}{2pt}
    \small{
        \begin{tabular}{cc}
            \begin{subtable}[t]{0.48\columnwidth}
                \centering
                \renewcommand{\arraystretch}{0.7}
                \begin{tabular}{l c}
                \toprule
                    \textbf{Component} & \textbf{Time} $\downarrow$ \\
                    & (ms) \\
                    \midrule
                    Image Encode & $20$ \\
                    Lift-contract-splat &  $14$  \\
                    BEV processing & $34$ \\
                    Decode $640$k queries & $18$ \\ \midrule
                     \textbf{Total} & $\mathbf{86}$ \\ 
                     \bottomrule
                \end{tabular}
                \label{tab:ablations:component-time}
            \end{subtable}
            &
            \begin{subtable}[t]{0.48\columnwidth}
                \renewcommand{\arraystretch}{0.7}
                \centering
                \begin{tabular}{lr}
                \toprule
                    \textbf{Model} & \textbf{FPS}$\uparrow$ \\ \midrule
                    \rowcolor{CornflowerBlue!10} SelfOcc & $1.2$ \\
                    \rowcolor{CornflowerBlue!10} OccNeRF & $1.3$ \\
                    \rowcolor{CornflowerBlue!10} GaussianOcc & $5.6$ \\
                    \rowcolor{CornflowerBlue!10} GaussTR & $0.2$ \\
                    \rowcolor{CornflowerBlue!10} GaussianFlowOcc & $10.2$ \\ 
                    \bestrow \textbf{\modelname} & $\mathbf{11.6}$\\
                    \bottomrule
                \end{tabular}
                \label{tab:ablations:inference-speed}
            \end{subtable}
            \\
        \end{tabular}
    }
    \caption{Per-component runtimes and frame-rate comparison. 
    \modelname reaches 11.6 FPS on an A100.
    \colorbox{CornflowerBlue!25}{FPS from \cite{boeder2025gaussianflowocc} on A100.}}
    \label{tab:efficiency}
\end{table}

\begin{table*}[t]
\centering
\setlength{\tabcolsep}{1pt}
\renewcommand{\arraystretch}{0.7}

\newcommand{\clsname}[2]{%
  \rotatebox{70}{\textcolor{#2}{$\blacksquare$}~\textsf{\scriptsize #1}}%
}
\begin{tabular}{
  l
  c
  *{15}{c}
}
\toprule
\textbf{Model} & \textbf{Mean} &
\clsname{barrier}{barrier} &
\clsname{bicycle}{bicycle} &
\clsname{bus}{bus} &
\clsname{car}{car} &
\clsname{cons. veh.}{construct} &
\clsname{drive. surf.}{driveable} &
\clsname{manmade}{manmade} &
\clsname{motorcycle}{motor} &
\clsname{pedestrian}{pedestrian} &
\clsname{sidewalk}{sidewalk} &
\clsname{terrain}{terrain} &
\clsname{traffic cone}{traffic} &
\clsname{trailer}{trailer} &
\clsname{truck}{truck} &
\clsname{vegetation}{vegetation} \\
\midrule
SelfOcc~\cite{huang2024selfocc} &
10.5 &
0.2 & 0.7 & 5.5 & 12.5 & 0.0 & \cellcolor{tabthird}55.5 & 14.2 & 0.8 & 2.1 & \cellcolor{tabthird}26.3 & \cellcolor{tabthird}26.5 & 0.0 & 0.0 & 8.3 & 5.6 \\

OccNeRF~\cite{zhang2023occnerf} &
10.8 &
0.8 & 0.8 & 5.1 & 12.5 & 3.5 & 52.6 & 18.5 & 0.2 & 3.1 & 20.8 & 24.8 & 1.8 & 0.5 & 3.9 & 13.2 \\

DistillNeRF~\cite{wang2024distillnerf} &
10.1 &
1.4 & 2.1 & 10.2 & 10.1 & 2.6 & 43.0 & 14.1 & 2.0 & 5.5 & 16.9 & 15.0 & 4.6 & 1.4 & 7.9 & 15.1 \\

LangOcc~\cite{boeder2025langocc}  &
13.3 &
3.1 & \cellcolor{tabsecond}9.0 & 6.3 & 14.2 & 0.4 & 43.7 & \cellcolor{tabthird}19.6 & \cellcolor{tabthird}10.8 & 6.2 & 9.5 & 26.4 & 9.0 & \cellcolor{tabsecond}3.8 & 10.7 & \cellcolor{tabfirst}26.4 \\

GaussianOcc~\cite{gan2025gaussianocc}   &
11.3 &
1.8 & 5.8 & 14.6 & 13.6 & 1.3 & 44.6 & 8.6 & 2.8 & \cellcolor{tabthird}8.0 & 20.1 & 17.6 & \cellcolor{tabthird}9.8 & 0.6 & 9.6 & 10.3 \\

GaussTR~\cite{jiang2025gausstr} {\footnotesize-FeatUp} &
13.3 &
2.1 & 5.2 & 14.1 & \cellcolor{tabthird}20.4 & \cellcolor{tabsecond}5.7 & 39.4 & \cellcolor{tabsecond}21.2 & 7.1 & 5.1 & 15.7 & 22.9 & 3.9 & 0.9 & \cellcolor{tabthird}13.4 & \cellcolor{tabthird}21.9 \\

GaussTR~\cite{jiang2025gausstr} {\footnotesize-T2D} &
\cellcolor{tabthird}13.9 &
\cellcolor{tabthird}6.5 & \cellcolor{tabthird}8.5 & \cellcolor{tabsecond}21.8 & \cellcolor{tabfirst}24.3 & \cellcolor{tabfirst}6.3 & 37.0 & \cellcolor{tabsecond}21.2 & \cellcolor{tabfirst}15.5 & 7.9 & 17.2 & 7.2 & 1.9 & \cellcolor{tabfirst}6.1 & \cellcolor{tabsecond}17.2 & 10.0 \\

GaussianFlowOcc~\cite{boeder2025gaussianflowocc} &
\cellcolor{tabsecond}17.1 &
\cellcolor{tabsecond}7.2 & \cellcolor{tabfirst}9.3 & \cellcolor{tabthird}17.6 & 17.9 & 4.5 & \cellcolor{tabsecond}63.9 & 14.6 & 9.3 & \cellcolor{tabsecond}8.5 & \cellcolor{tabsecond}31.1 & \cellcolor{tabsecond}35.1 & \cellcolor{tabsecond}10.7 & 2.0 & 11.8 & 12.6 \\

\textbf{\modelname} &
\cellcolor{tabfirst}21.3 &
\cellcolor{tabfirst}7.3 & 6.8 & \cellcolor{tabfirst}26.5 & \cellcolor{tabsecond}20.9 & \cellcolor{tabthird}4.8 & \cellcolor{tabfirst}69.2 & \cellcolor{tabfirst}25.2 & \cellcolor{tabsecond}10.9 & \cellcolor{tabfirst}15.0 & \cellcolor{tabfirst}34.5 & \cellcolor{tabfirst}38.4 & \cellcolor{tabfirst}13.2 & \cellcolor{tabthird}3.7 & \cellcolor{tabfirst}17.3 & \cellcolor{tabsecond}25.7 \\ \midrule
\textbf{\modelnameplus} & 23.5 & 9.0 & 10.0 & 30.4 & 25.5 &	4.6 & 69.6 & 28.0 &	16.5 & 17.0 & 37.2 &	42.4 &	11.8 &	3.4 & 18.5 & 28.8 \\
\bottomrule
\end{tabular}
\caption{Per-class \textbf{IoU}$\uparrow$ performance on the Occ3D-nuScenes dataset. \modelname performs better than prior work on many classes except the most rare ones. 
\modelnameplus uses high-res images as input, camera and lidar for supervision, and VFM feature distillation supervision, pushing the limits for our framework and performs best of all methods. Colors indicate \colorbox{tabfirst!75}{First}, \colorbox{tabsecond!75}{Second}, \colorbox{tabthird!75}{Third} in ranking.}
\label{tab:experiments:classwise-results-iou}
\end{table*}

\begin{table}[t]
\centering
\setlength{\tabcolsep}{1pt}
\renewcommand{\arraystretch}{0.7}
\begin{tabular}{
        C{1.5cm} C{1.5cm} C{1.4cm} C{1.5cm} |
        C{0.7cm} C{0.7cm} C{0.7cm}
    }
    \toprule
    \multicolumn{4}{c|}{\textbf{Components}} &
    \multicolumn{3}{c}{\textbf{RayIoU}$\uparrow$} \\
    \cmidrule(lr){1-4}\cmidrule(lr){5-7}
    Log-Lin. & Long Sup. & Contract & Point Enc. & \semantic & \dynamic & \occupancy \\
    \midrule
    \rowcolor{Apricot!50} & &  && 21.7 & 18.7 & 43.5  \\
    \cmark  &  & &  & 21.7 & 19.0 & 43.6\\
    \cmark &\cmark  & & & 22.9 & 20.8 & 44.7 \\
    \cmark & \cmark & \cmark & & 23.1 & 21.1 & 44.9 \\
     & \cmark  & \cmark  & & 22.5 & 20.3 & 44.3 \\
     & \cmark  & \cmark  & \cmark & 22.9 & 20.4 & 44.9 \\
    \bestrow
    \cmark & \cmark & \cmark  & \cmark & \textbf{23.6} & \textbf{21.7} & \textbf{45.2} \\
    \bottomrule
    \end{tabular}
    \caption{All proposed components individually improve performance and the full model performs better than the original \colorbox{Apricot!50}{LSS}.}
    \label{tab:ablations:lift-contract-splat}
\end{table}

\subsection{Main results}
\label{sec:experiments:main}
\modelname sets a new state-of-the-art (SOTA) among self-supervised camera-based methods on Occ3D-nuScenes, surpassing all prior approaches across both semantic and geometric metrics (\cref{tab:main_results}).
Compared to the strongest baseline, GaussianFlowOcc, it improves semantic RayIoU by \textbf{+26\%} and occupancy RayIoU by \textbf{+25\%}, while maintaining real-time inference at 11.6 FPS and a total latency of 86 ms (\cref{tab:efficiency}).
Notably, \modelname also outperforms GaussTR, despite GaussTR using higher-resolution inputs and vision-foundation-models at inference.
We note that \modelname is robust to choice of image encoder and shows SOTA performance also using \eg a ResNet-50 backbone in \supmat \cref{sec:supmat:experiment-details}.
Class-wise results in \cref{tab:experiments:classwise-results-iou} shows that QueryOcc particularly excels in small and thin categories, such as traffic cones and pedestrians, while also showing strong performance on large categories, \eg, drivable surface, sidewalk, and terrain. 
However, it is challenged by rare categories such as bicycle ($0.03\%$ of the data).
Similar trends are also reflected in RayIoU, as shown in the \supmat \cref{tab:supmat:classwise-results-rayiou}.

We believe these performance gains and efficiency stem from two core design choices: 
(1) spatio-temporal query supervision, which supplies direct supervision, and 
(2) a contractive BEV representation with explicitly encoded occupancy and occlusion encoding, which preserves near-field detail while extending supervision range without quadratic memory growth, and allows for long range reasoning.

The enhanced variant \modelnameplus uses high resolution images, combines camera–lidar supervision, and VFM feature distillation, pushes performance even further, achieving the best overall results across all metrics.
Overall, \modelname establishes a new benchmark for self-supervised 3D scene understanding from cameras, achieving high geometric accuracy, high efficiency, and strong performance on fine-scale structures.
Qualitative examples shown in \supmat \cref{sec:supmat:qualitative:examples}.

\subsection{Ablations}
We conduct controlled ablations to isolate how model design and supervision strategy contribute to performance. 

\parsection{Supervision signals}
We compare two forms of supervision within our framework: query-based 3D supervision and a rendering-based alternative that uses identical supervision sources (\ie metric depth, semantic pseudo-labels, and VFM features) but applies losses in image space via alpha blending (\cref{sec:supmat:supervision}).
\Cref{tab:ablations:supervision} shows that rendering supervision performs markedly worse (15.0 vs.\ 23.6 mRayIoU) in our framework, indicating that image-space reconstruction provides weaker geometric constraints.
Direct occupancy and semantic query supervision yield better performance, and adding VFM feature distillation offers further gains.
Combining rendering and query-based losses brings no benefit, implying that once strong 3D supervision is present, 2D losses are redundant.
Overall, we believe query-based supervision provides clearer geometric feedback and more reliable training signals for 3D scene learning.

\parsection{Model components}
As \cref{tab:ablations:lift-contract-splat} shows, log-linear depth bins, long supervision range, spatial contraction, and point encoding all contribute to performance gains.
These components jointly allow the BEV features to represent free, occupied, and occluded regions more distinctly at unbounded ranges.
The learned representations in \cref{fig:contraction-features} exemplifies this behavior, showing clear spatial organization of free, occupied and occluded regions.
Note that for all experiments we keep the memory and the compute the same, \ie a fixed amount of cells in the BEV grids. 
Short-range variants (Row 1 \& 2) have finer spatial resolution and far away queries (beyond 40\,m range) are excluded.
For long-range without contraction (Row 3), far away queries/features are mapped to the BEV boundary.

\begin{table}[t]
\centering
\setlength{\tabcolsep}{2pt}
\renewcommand{\arraystretch}{0.7}
\begin{tabular}{
    L{2.8cm} |
    *{3}{C{0.7cm}} |
    *{3}{C{0.7cm}}
}
\toprule
\multirow{2}{*}{\textbf{Setting}} &
\multicolumn{3}{c|}{\textbf{Supervision}} &
\multicolumn{3}{c}{\textbf{RayIoU}$\uparrow$} \\ 
\cmidrule(lr){2-4}\cmidrule(lr){5-7}
& Occ. & Sem. & Feat. & \semantic & \dynamic & \occupancy \\
\midrule
Rendering only & \cmark & \cmark & \cmark & 15.0 & 9.8 & 41.7 \\
\midrule
Query-based & \cmark & \cmark & & 23.6 & 21.7 & 45.2 \\
\bestrow Query-based +Feat. & \cmark & \cmark & \cmark & \textbf{24.0} & \textbf{22.0} & \textbf{45.7} \\
\midrule
Query + Rendering & \cmark & \cmark & \cmark & 23.3 & 21.3 & 45.6 \\
\bottomrule
\end{tabular}
\caption{Different supervision signals. 
Rendering supervision performs worse than direct query-based self-supervision. 
Adding feature supervision improves query-based training, while combining rendering and query-based shows no benefits.}
\vspace{6pt}
\label{tab:ablations:supervision}
\end{table}

\begin{figure*}[t]
\centering
\setlength{\tabcolsep}{1pt}
\renewcommand{\arraystretch}{0.7}

\begin{minipage}[t]{0.32\textwidth}
  \vspace{0pt} 
  \centering
  \begin{tabular}{
      C{1.1cm} C{1.1cm} |
      C{0.7cm} C{0.7cm} C{0.7cm}
  }
  \toprule
  \multicolumn{2}{c|}{\textbf{PCD Source}} &
  \multicolumn{3}{c}{\textbf{RayIoU}$\uparrow$} \\
  \cmidrule(lr){1-2}\cmidrule(lr){3-5}
  Pseudo & Lidar & \semantic & \dynamic & \occupancy \\
  \midrule
  \cmark &        & 23.6 & 21.7 & 45.2 \\
         & \cmark & 22.9 & 19.4 & 48.3 \\
  \bestrow
  \cmark & \cmark & \textbf{24.6} & \textbf{22.5} & \textbf{48.7} \\
  \bottomrule
  \end{tabular}
   \vspace{4pt}
  \caption{\modelname learns effectively from different point cloud sources. Combining both improves performance.}
  \label{tab:ablations:point_cloud_source}
\end{minipage}
\hfill
\begin{minipage}[t]{0.32\textwidth}
  \vspace{-6pt}
  \centering
  \input{figures/pseudopcd_subsampling.tex}
   \vspace{-16pt}
  \caption{Effect of pseudo point cloud subsampling. Stable performance with fewer than $\sim$100k points.}
  \label{fig:pseudopcd-subsampling}
\end{minipage}
\hfill
\begin{minipage}[t]{0.32\textwidth}
  \vspace{-8pt}
  \centering
  \input{figures/image_resolution_scaling}
  \vspace{-2pt}
  \caption{Impact of input image resolution. Accuracy improves with scale, demonstrating model efficiency.}
  \label{fig:image-resolution}
\end{minipage}
\end{figure*}

\subsection{Framework scalability and data efficiency}
\label{sec:framework-scalability}
A scalable self-supervised framework must adapt to different supervision sources, handle dense data efficiently, and continue improving as input fidelity and dataset size increase.
We therefore evaluate four aspects of scalability: (1) supervision modality (camera-only vs.\ camera-\lidar), (2) query sampling efficiency, (3) input image resolution, and (4) dataset scale.
Together, these experiments assess the framework's ability to generalize across sensor setups, leverage richer inputs, and benefit from more data without architectural or supervision scheme changes.

\parsection{Supervision source}
We compare training using different \anypcd sources: \pseudopcd generated from camera images via a vision foundation model, and \lidarpcd.
\Cref{tab:ablations:point_cloud_source} summarizes the results.
Pseudo-supervision performs better on semantic metrics, potentially due to its higher point density and perfect temporal alignment with the camera.
In contrast, \lidar supervision achieves higher geometric accuracy (\occupancy) thanks to its precise depth measurements.
We hypothesize that the lower \semantic~performance is due to projection and time synchronization errors.
Combining both sources yields the best overall results, showing that the framework can effectively fuse dense semantic cues with accurate geometry.
This flexibility enables training on datasets with different sensor configurations without modifying the architecture or loss formulation.

\parsection{Query sampling}
The pseudo point clouds generated by the vision foundation model are extremely dense, containing about $1.4$M points per image for nuScenes~\cite{caesar2020nuscenes}.
Projecting every pixel across all cameras and time steps is computationally costly, so we analyze how subsampling affects performance and efficiency.
As shown in \cref{fig:pseudopcd-subsampling}, performance remains nearly unchanged when sub-sampling points per frame, but reduces training time substantially when using fewer than $100$k points.
This demonstrates that the framework remains data-efficient even with sparse 3D supervision.
Alternative sampling strategies are investigated in \supmat \cref{sec:supmat:results}.

\parsection{Increasing image resolution}
We evaluate how input resolution affects performance while keeping architecture and supervision fixed. 
As shown in \cref{fig:image-resolution}, both the main and VFM-supervised models improve steadily with higher resolution, gaining nearly +3 mRayIoU between $256\times256$ and $900\times1600$. 
The consistent gap between the two variants indicates that feature-distillation supervision provides complementary cues independent of input scale.
The framework can exploit additional image detail to increase its geometric understanding, demonstrating that representation quality continues to improve with richer inputs alone. 

\parsection{Additional data}
We evaluate scalability by adding Argoverse 2~\cite{Argoverse2} training data, using \lidar-derived pseudo point clouds for supervision. 
As shown in \Cref{tab:datasets-pseudo}, incorporating AV2 substantially improves performance across all metrics, increasing semantic RayIoU from 22.9 to 26.0 and dynamic RayIoU from 19.4 to 23.8.
These gains demonstrate that \modelname effectively leverages heterogeneous data sources without requiring explicit domain adaptation or label harmonization. 
This suggests that the learned continuous 4D representation generalizes across dataset-specific biases in sensor setup and scene distribution.
Further scaling with additional data using only VFM-based supervision yields consistent improvements (see \supmat \cref{sec:supmat:datasetsize}), indicating that the framework benefits from increased data diversity irrespective of the supervision modality.

\begin{table}
    \centering
    \setlength{\tabcolsep}{3pt} 
    \renewcommand{\arraystretch}{0.7} 
    \vspace{2pt}
    \begin{tabular}{
        C{0.7cm} C{0.7cm}  
        C{0.9cm} C{0.9cm} C{0.9cm}  
    }
    \toprule
    \multicolumn{2}{c}{\textbf{Datasets}} & \multicolumn{3}{c}{\textbf{RayIoU}$\uparrow$} \\
    \cmidrule(lr){3-5} \cmidrule(lr){1-2}
    nuSc & AV2 & \semantic & \dynamic & \occupancy \\
    \midrule
    \cmark &                  & 22.9 & 19.4 & 48.3 \\
    \cmark & \cmark           & 26.0 & 23.8 & 50.8 \\
    \bottomrule
    \end{tabular}
    \caption{Adding AV2 data improves the performance. The point cloud source is lidar.}
    \vspace{-4pt}
    \label{tab:datasets-pseudo}
\end{table}

%% file: figures/pseudopcd_subsampling.tex
\pgfplotstableread{
Name mRayIoU
5	23.2
25	23.4
50	23.3
100	23.2
200	23.3
300	23.5
}\resultstable
\pgfplotstableread{
Name time
5	12.8
25	13.1
50	13.1
100	13.1
200	16.5
300	20.3
}\timestable

\begin{tikzpicture}
\begin{axis}[
  width=\columnwidth,
  height=\columnwidth/2,
  xlabel={Sampled points (in thousands, log)},
  ylabel={mRayIoU},
  ylabel style={yshift=-6pt},
  xmin=0, xmax=310,
  xmode=log,
  log basis x=10,
  xtick={5,25,50,100,300},
  xticklabels={5,25,50,100,300},
  xticklabel style={/pgf/number format/fixed},
  ymin=18, ymax=26,
  grid=both,
  grid style={dotted,gray!30},
  every axis/.append style={
    tick label style={font=\footnotesize},
    label style={font=\footnotesize},
    title style={font=\footnotesize}
  },
  legend style={
    font=\footnotesize,
    at={(0.25,1.05)},
    anchor=south,
    legend columns=-1,
    draw=none
  },
  legend cell align=left,
  every axis plot/.append style={thick,mark size=2.5pt},
]

\addplot+[
  mark=*,
  color={rgb,255:red,33; green,113; blue,181},
  solid,
  line width=0.8pt,        
  mark size=1.8pt,         
  mark options={
    solid,
    fill={rgb,255:red,33; green,113; blue,181},
    rotate=0,              
    line width=0.3pt       
  }
]
      table[y=mRayIoU, x=Name]
      {\resultstable};
    \addlegendentry{RayIoU}

  \end{axis}

\begin{axis}[
  width=\columnwidth,
  height=\columnwidth/2,
  axis y line*=right,
  axis x line=none,
  ylabel={Training time (h)},
  ylabel style={yshift=6pt},
  ymin=10, ymax=22,
  xmin=0, xmax=310,
  xmode=log,
  log basis x=10,
  xtick={5,25,50,100,300},
  xticklabels={5,25,50,100,300},
  xticklabel style={/pgf/number format/fixed},
  every axis/.append style={
    tick label style={font=\footnotesize},
    label style={font=\footnotesize},
    title style={font=\footnotesize}
  },
  legend style={
    font=\footnotesize,
    at={(0.75,1.05)},
    anchor=south,
    legend columns=-1,
    draw=none
  },
  legend cell align=left,
]

\addplot+[
  mark=square*,
  color={rgb,255:red,33; green,113; blue,181},
  dashed,
  line width=0.8pt,        
  mark size=1.8pt,         
  mark options={
    solid,
    fill={rgb,255:red,33; green,113; blue,181},
    rotate=0,              
    line width=0.3pt       
  }
]
table[y=time, x=Name] {\timestable};
    \addlegendentry{Training time}
  \end{axis}

\end{tikzpicture}

%% file: figures/image_resolution_scaling.tex
\pgfplotstableread{
Supervision Resolution mRayIoU bRayIoU
Main 256x256 21.6 43.4
Main 256x704 23.2 44.5
Main 504x896 23.9 45.6
Main 900x1600 24.7 45.9
}\maintable
\pgfplotstableread{
Supervision Resolution mRayIoU bRayIoU
VFM 256x256 21.9 43.5
VFM 256x704 23.9 45.0
VFM 504x896 25.0 46.2
VFM 900x1600 25.8 46.4
}\vfmtable

\begin{tikzpicture}
  \begin{axis}[
    width=\columnwidth,
    height=\columnwidth/2,
    xlabel={Input resolution},
    ylabel={mRayIoU},
    ylabel style={yshift=-6pt},
    symbolic x coords={256x256,256x704,504x896,900x1600},
    xtick=data,
    x tick label style={rotate=00},
    ymin=21, ymax=27,
    grid=both,
    grid style={dotted,gray!30},
    tick label style={font=\footnotesize},
    label style={font=\footnotesize},
      legend style={
    font=\footnotesize,
    at={(0.5,1.05)},
    anchor=south,
    legend columns=-1,
    draw=none            
  },
  legend cell align=left,
    every axis plot/.append style={thick,mark size=1.8pt},
  ]
    \addplot+[mark=*,
      color={rgb,255:red,33; green,113; blue,181},
      mark options={fill={rgb,255:red,33; green,113; blue,181}}]
      table[y=mRayIoU, x=Resolution]
      {\maintable};
    \addlegendentry{Main }

    \addplot+[mark=triangle*,
      color={rgb,255:red,57; green,181; blue,74},
      mark options={fill={rgb,255:red,57; green,181; blue,74}}]
      table[y=mRayIoU, x=Resolution]
      {\vfmtable};
    \addlegendentry{w. VFM}
  \end{axis}


\end{tikzpicture}

%% file: sec/5_conclusions.tex
\section{Conclusion}
\label{sec:conclusion}
We introduced \modelname, a self-supervised framework that learns a continuous 3D semantic occupancy field from multi-view images.
It achieves state-of-the-art performance on Occ3D-nuScenes among self-supervised methods.
This improvement stems from the proposed query-based supervision scheme, which directly supervises geometry and semantics through spatio-temporal 4D queries across adjacent frames, while eliminating the need for rendering losses or voxelized lidar aggregation.
The proposed contractive BEV representation enables efficient computation and real-time inference while preserving geometric detail and long-range supervision in unbounded scenes.
\modelname flexibly supports both camera-only supervision via pseudo point clouds and camera–lidar setups, adapting to available sensor configurations.
Performance scales consistently with higher image resolution and additional datasets, indicating strong data efficiency and cross-domain generalization.
Together, these results establish direct 4D query supervision as a robust foundation for large-scale self-supervised 3D learning.

\parsection{Limitations and future work}
Fundamentally, all current approaches are limited by being supervised by what is observed by the onboard sensors, and thus struggle to represent occluded parts of the scene. 
Future work could address this by incorporating representation learning objectives that give supervision signals in unobservable regions through consistency or feature reconstruction.
Another promising direction is collaborative supervision across multiple agents, where overlapping viewpoints from different vehicles or time-shifted observations provide complementary supervision signals, \eg, enabling the model to \textit{see} around corners.

%% file: sec/6_acknowledgements.tex
\vspace{-4mm}
\subsection*{Acknowledgements}
We thank Carl Lindström, William Ljungbergh, Georg Hess, and Adam Tonderski for fruitful discussions and valuable feedback on the manuscript. 
This work was partially supported by the Wallenberg AI, Autonomous Systems and Software Program (WASP) funded by the Knut and Alice Wallenberg Foundation.
Computational resources were provided by NAISS at \href{https://www.nsc.liu.se/}{NSC Berzelius} and \href{https://www.c3se.chalmers.se/about/Alvis/}{C3SE Alvis}, partially funded by the Swedish Research Council, grant agreement no. 2022-06725.

%% file: sec/X_suppl.tex
\clearpage
\setcounter{page}{1}

\maketitlesupplementary
\section*{Overview of Supplementary Material}

This document provides additional experimental details, analysis, and qualitative results that complement the main paper.
It includes:
\begin{itemize}
    \item Dataset and class details used for supervision and evaluation (\cref{sec:supmat:dataset}). 
    \item Implementation and training specifics for the proposed method and rendering supervision scheme (\cref{sec:supmat:experiment-details}, \cref{sec:supmat:supervision}). 
    \item Complete per-class metrics for RayIoU and additional experiments (\cref{sec:supmat:results}).
    \item Qualitative examples, including long-range predictions and behavior in contracted regions (\cref{sec:supmat:qualitative:examples}).
    \item Discussion of the discrepancy between continuous predictions and voxelized ground truth (\cref{sec:supmat:discussion}).
\end{itemize}

\begin{table}[b]
\centering
\setlength{\tabcolsep}{3pt}
\renewcommand{\arraystretch}{0.9}

\newcommand{\clsname}[2]{%
  \rotatebox{0}{\textcolor{#2}{$\blacksquare$}~#1}%
}

\begin{tabular}{L{0.4cm} L{2.7cm} L{4.7cm}}
\toprule
\textbf{} & \textbf{Class} & \textbf{VFM Prompt Vocabulary} \\
\midrule

\multirow{7}{*}{\rotatebox{90}{\textbf{Static classes}}}
 & \clsname{barrier}{barrier} & barricade, barrier \\
 & \clsname{traffic cone}{traffic} & traffic-cone \\
 & \clsname{sidewalk}{sidewalk} & sidewalk, walkway \\
 & \clsname{drive. surf.}{driveable} & highway, street \\
 & \clsname{terrain}{terrain} & grass, sand, gravel, terrain \\
 & \clsname{vegetation}{vegetation} & bush, plants, tree \\
 & \clsname{manmade}{manmade} & building, wall, fence, pole, sign, light, bridge, billboard \\
\midrule

\multirow{8}{*}{\rotatebox{90}{\textbf{Dynamic classes}}}
 & \clsname{bicycle}{bicycle} & bicycle \\
 & \clsname{motorcycle}{motor} & motorcycle, scooter \\
 & \clsname{cons. veh.}{construct} & excavator, crane \\
 & \clsname{pedestrian}{pedestrian} & person, pedestrian \\
 & \clsname{trailer}{trailer} & trailer \\
 & \clsname{truck}{truck} & lorry, truck, tractor \\
 & \clsname{bus}{bus} & bus \\
 & \clsname{car}{car} & car, vehicle, sedan, van, jeep \\
\bottomrule
\end{tabular}

\caption{Semantic classes used for training and evaluation, grouped into static and dynamic categories and their corresponding prompt vocabularies used for pseudo-label generation.}
\label{tab:dataset:classes}
\end{table}

\section{Dataset details}
\label{sec:supmat:dataset}
We follow prior work~\cite{jiang2025gausstr, boeder2025gaussianflowocc} and focus on 15 semantic classes from the Panoptic nuScenes dataset~\cite{fong2022panoptic}. 
These classes are grouped into \emph{static} and \emph{dynamic} categories according to their motion characteristics, as summarized in \cref{tab:dataset:classes}.
As \cref{tab:supmat:classwise-frequencies} shows, the dataset is a highly class-imbalanced and that the class distributions between lidar labels and pseudo-labels from images varies. 
For instance, the lidar point clouds on average consist of $38\%$ drivable surface, whereas the image-plane pseudo labels are $22\%$ in the same category.

\section{Experiment Details}
\label{sec:supmat:experiment-details}
\parsection{Image encoder}
For the image encoder, we adopt ConvNeXt-Base~\cite{convnext} pretrained with DINOv3~\cite{simeoni2025dinov3}. 
As \cref{tab:supmat:backbone_ablation} shows, \modelname achieves strong performance across a broad range of backbones, including lightweight ResNets, pretrained on ImageNet \cite{imagenet}, to larger ConvNeXt variants, indicating that the framework is not tied to a specific encoder architecture. 
ConvNeXt-Base offers the highest semantic and geometric accuracy while still maintaining real-time inference.
The results also highlight a predictable trade-off: larger encoders improve semantic occupancy but reduce throughput. 

\parsection{Lift-contract-splat}
The point encoder $F_\theta$ uses Fourier features with 16 log–linear frequency bands in the range $[1,10]$, followed by a lightweight MLP that mixes the spatial and temporal inputs. 
For lift–contract–splat, we set $\alpha = 0.3$, $d_{\text{near}} = 40$\,m, and $d_{\text{far}} = 100$\,m, and include an 'infinity bin' at $180$\,m to cover the maximum sensor range. 
We supervise the categorical depth predictions, with weight $\lambda_{\text{depth}}$ in the total loss, using cross-entropy.
The VFM depth predictions and projected lidar points are used for the pseudo point cloud and real lidar point cloud trainings respectively.
The BEV grid maintains full resolution within $\pm40$\,m in $x$ and $y$, and contracts points outside this region using $\beta = 0.8$.
Effective cell side length within the full resolution region is $0.16$m.

\parsection{Core model}
The network uses a hidden dimension of $160$ throughout.
The BEV encoder consists of $10$ residual blocks and four multi-scale deformable-attention layers with pre-normalization. 
The entire model is implemented in pure PyTorch without custom CUDA kernels.

\parsection{Training}
Training runs for $300$k steps using AdamW (weight decay $10^{-4}$), batch size $4$, a peak learning rate of $10^{-6}$ for the image encoder and $10^{-4}$ for the remaining modules, with cosine decay and $6$k warmup steps. 
VFM feature distillation uses an $L_1$ loss. 
The losses are weighted: $\lambda_{\text{occ}} = 1.0$, $\lambda_{\text{sem}} = 0.5$, $\lambda_{\text{vfm}} = 0.5$, and $\lambda_{\text{depth}} = 0.5$.
We use log-frequency class weighting (using the frequencies in \cref{tab:supmat:classwise-frequencies} for each supervision source respectively) without any further tuning.
For the combined pseudo point cloud and real lidar point cloud trainings, we use the lidar frequencies.
We follow the same prompts, listed in \cref{tab:dataset:classes}, for generating pseudo-labels from GroundedSAM \cite{ren2024grounded} as GaussianFlowOcc \cite{boeder2025gaussianflowocc} for fair comparison.
We supervise $3$ frames forward and $3$ frames backward per sample and subsample $30$k 3D points per frame for efficiency. 
All training and inference benchmarks are obtained on A100 GPUs.

\begin{table}[t]
\centering
\setlength{\tabcolsep}{1pt}
\renewcommand{\arraystretch}{0.7}

\begin{tabular}{
    l
    C{0.7cm} C{0.7cm} C{0.7cm} 
    C{0.7cm} C{0.7cm} C{0.7cm} 
    C{0.7cm}                 
}
\toprule
\multirow{2}{*}{\textbf{Method}}
 & \multicolumn{3}{c}{\textbf{RayIoU}$\uparrow$}
 & \multicolumn{3}{c}{\textbf{IoU}$\uparrow$}
 & \multirow{2}{*}{\textbf{FPS}$\uparrow$} \\
\cmidrule(lr){2-4} \cmidrule(lr){5-7}
 & \semantic & \dynamic & \occupancy
 & \semantic & \dynamic & \occupancy
 &  \\
\midrule

GaussianFlowOcc                 & 18.7 & --   & --   & 17.1 & 10.1 & 46.9 & 10.2 \\
GaussianFlowOcc*                & 18.2 & 17.2 & 36.0 & 16.1 &  9.9 & 40.2 & 10.2 \\ \midrule

\rowcolor{gray!10} ResNet18                   & 20.4 & 17.4 & 42.0 & 18.3 & 9.7  & 51.6 & 14.1 \\
\rowcolor{gray!10} ResNet34                   & 20.8 & 17.8 & 43.1 & 19.0 & 9.9  & 53.3 & 13.8 \\
\rowcolor{gray!10} ResNet50                   & 22.0 & 19.3 & 43.7 & 19.6 & 11.1 & 52.0 & 13.2 \\
\rowcolor{gray!10} ResNet101                  & 22.5 & 20.4 & 43.9 & 20.5 & 12.3 & 53.9 & 12.5 \\

\rowcolor{gray!10} ConvNeXt-Tiny              & 22.2 & 19.7 & 43.9 & 20.0 & 11.3 & 54.0 & 12.9 \\
\rowcolor{gray!10} ConvNeXt-Small             & 22.8 & 20.6 & 44.3 & 20.5 & 12.3 & 54.1 & 12.0 \\
\rowcolor{gray!10} ConvNeXt-Base              & \textbf{23.6} & \textbf{21.7} & \textbf{45.2}
                                & \textbf{21.3} & \textbf{13.2} & \textbf{55.0}
                                & 11.6 \\
\bottomrule
\end{tabular}

\caption{Comparison of image encoders in our \colorbox{gray!10}{\modelname} framework. 
All configurations use identical training hyperparameters and supervision.
We note that \modelname shows state-of-the-art performance using a wide range of image encoders.
ConvNeXt-Base performs best, while retaining high inference speed.
* denotes reproduction for both RayIoU and IoU.
}
\vspace{12pt}
\label{tab:supmat:backbone_ablation}
\end{table}

\begin{table*}
\centering
\setlength{\tabcolsep}{2pt}
\renewcommand{\arraystretch}{1.05}

\newcommand{\clsname}[2]{%
  \rotatebox{70}{\textcolor{#2}{$\blacksquare$}~#1}%
}
\begin{tabular}{
  l
  *{15}{c}
}
\toprule
\textbf{Source} &
\clsname{barrier}{barrier} &
\clsname{bicycle}{bicycle} &
\clsname{bus}{bus} &
\clsname{car}{car} &
\clsname{cons. veh.}{construct} &
\clsname{drive. surf.}{driveable} &
\clsname{manmade}{manmade} &
\clsname{motorcycle}{motor} &
\clsname{pedestrian}{pedestrian} &
\clsname{sidewalk}{sidewalk} &
\clsname{terrain}{terrain} &
\clsname{traffic cone}{traffic} &
\clsname{trailer}{trailer} &
\clsname{truck}{truck} &
\clsname{vegetation}{vegetation} \\
\midrule
\Lidar &
1.11 & 0.02 & 0.55 & 4.65 & 0.19 & 37.93 & 21.30 & 0.05 & 0.28 & 8.37 & 8.24 & 0.09 & 0.66 & 1.93 & 14.63 \\

Pseudo &
2.12 & 0.03 & 0.71 & 6.89 & 0.13 & 21.88 & 35.31 & 0.04 & 0.35 & 6.04 & 7.31 & 0.11 & 0.08 & 1.86 & 17.15 \\
\bottomrule
\end{tabular}
\caption{Per-class point cloud frequency (\%) for \lidarpcd and \pseudopcd supervision sources.}
\vspace{16pt}
\label{tab:supmat:classwise-frequencies}
\end{table*}

\begin{table*}[!t]
\centering
\setlength{\tabcolsep}{1pt}
\renewcommand{\arraystretch}{0.7}

\newcommand{\clsname}[2]{%
  \rotatebox{70}{\textcolor{#2}{$\blacksquare$}~\textsf{\scriptsize #1}}%
}

\begin{tabular}{
  l
  c
  *{15}{c}
}
\toprule
\textbf{Model} & \textbf{Mean} &
\clsname{barrier}{barrier} &
\clsname{bicycle}{bicycle} &
\clsname{bus}{bus} &
\clsname{car}{car} &
\clsname{cons. veh.}{construct} &
\clsname{drive. surf.}{driveable} &
\clsname{manmade}{manmade} &
\clsname{motorcycle}{motor} &
\clsname{pedestrian}{pedestrian} &
\clsname{sidewalk}{sidewalk} &
\clsname{terrain}{terrain} &
\clsname{traffic cone}{traffic} &
\clsname{trailer}{trailer} &
\clsname{truck}{truck} &
\clsname{vegetation}{vegetation} \\
\midrule
SelfOcc~\cite{huang2024selfocc}  &
10.9 &
0.9 & 1.0 & 15.5 & 15.7 & 0.0 & \cellcolor{tabsecond}48.6 & 16.4 & 0.8 & 5.0 & \cellcolor{tabsecond}13.6 & \cellcolor{tabfirst}17.5 & 0.0 & 0.0 & 19.7 & 8.7 \\

LangOcc~\cite{boeder2025langocc} &
11.6 &
1.9 & 5.2 & 12.9 & 18.2 & 0.2 & 20.8 & \cellcolor{tabsecond}23.6 & 8.9 & 12.8 & 3.4 & \cellcolor{tabthird}12.3 & \cellcolor{tabsecond}13.3 & \cellcolor{tabthird}1.7 & 11.7 & \cellcolor{tabfirst}27.6 \\

GaussTR~\cite{jiang2025gausstr} {\footnotesize-FeatUp} &
13.8 &
4.2 & \cellcolor{tabthird}8.0 & 24.5 & 26.2 & \cellcolor{tabfirst}9.3 & 23.0 & 17.3 & 8.2 & 13.7 & 4.5 & 8.9 & \cellcolor{tabthird}10.2 & 0.4 & \cellcolor{tabthird}25.3 & \cellcolor{tabthird}23.2 \\

GaussTR~\cite{jiang2025gausstr} {\footnotesize-T2D} &
\cellcolor{tabthird}14.2 &
\cellcolor{tabthird}9.4 & \cellcolor{tabfirst}12.7 & \cellcolor{tabthird}35.1 & \cellcolor{tabthird}30.0 & \cellcolor{tabsecond}8.5 & 18.9 & 19.4 & \cellcolor{tabfirst}14.8 & \cellcolor{tabsecond}17.1 & 5.3 & 3.2 & 3.0 & \cellcolor{tabfirst}4.3 & 19.2 & 12.4 \\

GaussianFlowOcc~\cite{boeder2025gaussianflowocc}  &
\cellcolor{tabsecond}18.2 &
\cellcolor{tabsecond}13.1 & \cellcolor{tabsecond}9.6 & \cellcolor{tabsecond}36.9 & \cellcolor{tabsecond}30.7 & 5.1 & \cellcolor{tabthird}39.8 & \cellcolor{tabthird}22.9 & \cellcolor{tabsecond}10.5 & \cellcolor{tabthird}15.7 & \cellcolor{tabthird}13.3 & \cellcolor{tabsecond}15.2 & \cellcolor{tabsecond}13.3 & 1.3 & \cellcolor{tabsecond}27.7 & 19.6 \\

\textbf{\modelname} &
\cellcolor{tabfirst}23.6 &
\cellcolor{tabfirst}13.6 & 6.2 & \cellcolor{tabfirst}49.1 & \cellcolor{tabfirst}33.9 & \cellcolor{tabthird}6.3 & \cellcolor{tabfirst}53.1 & \cellcolor{tabfirst}32.2 & \cellcolor{tabthird}9.4 & \cellcolor{tabfirst}26.1 & \cellcolor{tabfirst}19.2 & \cellcolor{tabfirst}17.5 & \cellcolor{tabfirst}19.3 & \cellcolor{tabsecond}2.5 & \cellcolor{tabfirst}39.9 & \cellcolor{tabsecond}25.7 \\ \midrule
\textbf{\modelnameplus} & 25.8 & 15.7 &	10.9 &	52.5 &	36.2 &	5.9 &	55.2 &	35.5 &	13.6 &	29.5 &	22.1 &	20.1 &	20.0 &	0.6 &	41.0 &	28.5 \\
\bottomrule
\end{tabular}
\caption{Per-class \textbf{RayIoU}$\uparrow$ performance on the Occ3D-nuScenes dataset.  \modelname performs better than prior work on all classes . 
\modelnameplus use high-res images, camera and lidar for supervision, and VFM feature distillation supervision pushing the limits for our framework and performs best of all methods. All baselines are reproduced results, and colors indicate \colorbox{tabfirst!75}{First}, \colorbox{tabsecond!75}{Second}, \colorbox{tabthird!75}{Third} in ranking.}
\vspace{12pt}
\label{tab:supmat:classwise-results-rayiou}
\end{table*}

\section{Rendering Supervision Details}
\label{sec:supmat:supervision}
To compare image-space 2D supervision and 4D query-based supervision in the same framework, we implement a rendering-based baseline following neural rendering formulations adapted to occupancy prediction \cite{huang2024selfocc, wang2024distillnerf}. 
The goal is to test 2D supervision pipeline within the same architecture and using identical supervision sources.

For each camera pixel, we cast a ray using known intrinsics and extrinsics, sample points along the ray with exponentially increasing depth spacing, and query the model's predicted occupancy $\hat{o}$ and feature vector $\hat{\mathbf{v}}$ at each location. 
Rendered features are obtained via alpha compositing:
\begin{equation}
    \hat{f}_{\text{rendered}}
    = \sum_{i} T_i \, o_i \, f_i,
    \qquad
    T_i = \prod_{j < i} (1 - o_j),
\end{equation}
where $\hat{o}$ acts as opacity and $T_i$ is accumulated transmittance. 
A lightweight 6-layer MLP (hidden dimension 64) predicts per-pixel semantic logits from the rendered features. 
We supervise rendered depth, semantics, and VFM features using the same supervision sources as in the main \modelname\ model: Metric3D depth, Grounded-SAM semantics, and DinoV3 features.

To mitigate depth ambiguity, we apply a single round of importance sampling around the predicted depth, concentrating samples near likely surface intersections. 

\section{Additional Results}
\label{sec:supmat:results}
This section describes additional experiments and results.

\subsection{Class-wise metrics}
We observe similar trends as for RayIoU in the IoU metrics shown in \cref{tab:supmat:classwise-results-rayiou}.
Drivable surface, sidewalk, and terrain achieve particularly high scores, suggesting that the model captures the ground-plane structure reliably.
\modelname also shows clear improvements on thin or small objects such as pedestrian and traffic cone, which can risk disappearing or blurring in BEV grids. 
The contracted BEV representation and query-based decoder appear to preserve local spatial detail even under noisy pseudo-labels.
Rare classes such as bicycle, trailer, and construction vehicle remain difficult for all methods due to severe class imbalance and possibly label noise. 
\modelname nevertheless avoids the collapse observed in several baselines and maintains competitive performance across these categories.

\subsection{Input image resolution}
We evaluate the effect of input resolution supervising both with and without vision foundation model (VFM) features in \cref{tab:resolution_ablation}.
Performance increases steadily with higher resolutions for both supervision settings, indicating that the framework can make use of additional image detail when available. 
As expected, higher resolutions reduce throughput, reflecting the quadratic cost of encoding and lifting denser feature maps. 
The consistent gap between the \emph{main} and \emph{w. VFM} settings show that feature-distillation supervision provides complementary cues independent of resolution.

\begin{table}[t]
\centering
\setlength{\tabcolsep}{1pt} 
\renewcommand{\arraystretch}{0.7}

\begin{tabular}{
    l
    c
    C{0.7cm} C{0.7cm} C{0.7cm}  
    C{0.7cm} C{0.7cm} C{0.7cm}  
    r                     
}
\toprule
\multirow{2}{*}{\textbf{Model}}
 & \multirow{2}{*}{\textbf{Resolution}}
 & \multicolumn{3}{c}{\textbf{RayIoU}$\uparrow$}
 & \multicolumn{3}{c}{\textbf{IoU}$\uparrow$}
 & \multirow{2}{*}{\textbf{FPS}$\uparrow$} \\
\cmidrule(lr){3-5} \cmidrule(lr){6-8}
 &  & \semantic & \dynamic & \occupancy
    & \semantic & \dynamic & \occupancy
    &  \\
\midrule

\multirow{4}{*}{Main} & 256$\times$256   & 21.6 & 18.9 & 43.4 & 19.1 & 10.2 & 54.1 & 14.1 \\
     & 256$\times$704   & 23.6 & 21.7 & 45.2 & 21.3 & 13.2 & 55.0 & 11.3 \\
     & 504$\times$896   & 23.9 & 22.0 & 45.6 & 21.9 & 14.0 & 55.1 &  6.4 \\
     & 900$\times$1600  & 24.7 & 23.1 & 45.9 & 22.9 & 15.3 & 55.5 &  2.4 \\
\midrule
\multirow{4}{*}{w. VFM}  & 256$\times$256   & 21.9 & 19.7 & 43.5 & 19.6 & 10.8 & 54.5 & 13.4 \\
     & 256$\times$704   & 23.9 & 22.3 & 45.0 & 21.5 & 13.5 & 55.5 & 10.9 \\
     & 504$\times$896   & 25.0 & 23.5 & 46.2 & 22.7 & 15.0 & 56.0 &  6.2 \\
     & 900$\times$1600  & 25.8 & 24.2 & 46.4 & 23.1 & 16.4 & 56.3 &  2.4 \\
\bottomrule
\end{tabular}

\caption{Effect of input resolution under the \emph{main} settings with semantic and occupancy supervision as well as \emph{with VFM} feature supervision. 
As the input resolution increases the performance goes up, but the inference speed goes down.
The VFM model is slightly slower due to the higher output dimension in the decoder head.
}
\vspace{12pt}
\label{tab:resolution_ablation}
\end{table}

\subsection{Pseudo-label source}
We compare different semantic pseudo-label sources in \cref{tab:samversion_ablation}.
GroundedSAMv1 \cite{ren2024grounded} is used in the main experiments to match prior work, but GroundedSAMv2 \cite{ren2024grounded} provides slightly higher-quality masks and improves semantic metrics.
The overall performance gap is small, which indicates that the framework is not overly sensitive to the specific pseudo-label source.

\begin{table}[t]
\centering
\setlength{\tabcolsep}{2pt} 
\renewcommand{\arraystretch}{0.7}

\begin{tabular}{
    l
    C{0.7cm} C{0.7cm} C{0.7cm}  
    C{0.7cm} C{0.7cm} C{0.7cm}  
}
\toprule
\multirow{2}{*}{\textbf{Pseudo Labels}}
 & \multicolumn{3}{c}{\textbf{RayIoU}$\uparrow$}
 & \multicolumn{3}{c}{\textbf{IoU}$\uparrow$} \\
\cmidrule(lr){2-4} \cmidrule(lr){5-7}
 & \semantic & \dynamic & \occupancy
 & \semantic & \dynamic & \occupancy \\
\midrule

GroundedSAM v1 & 23.6 & 21.7 & 45.2 & 21.3 & 13.2 & 55.0 \\
GroundedSAM v2 & 24.0 & 22.1 & 45.5 & 21.7 & 13.8 & 55.2 \\
\bottomrule
\end{tabular}

\caption{Effect of different pseudo-label sources. Using GroundedSAMv2 gives a performance boost on the semantic metrics.}
\vspace{6pt}
\label{tab:samversion_ablation}
\end{table}

\subsection{Pseudo-label noise}
We evaluate robustness to noisy supervision in \cref{tab:supmat:pseudo-label-noise} by injecting controlled perturbations into geometry and semantics. 
Depth noise is modeled as additive Gaussian noise on Metric3D's predicted inverse depth, while semantic noise is simulated by randomly flipping a fraction of pixel labels.
Performance degrades gracefully as noise increases, demonstrating that \modelname is robust to this type of supervision error.
We expect that as vision models continue to improve in accuracy and robustness, \modelname will naturally benefit from these advances.

\begin{table*}[h]
\centering
\setlength{\tabcolsep}{3pt}
\renewcommand{\arraystretch}{0.7}

\vspace{-1mm}
\begin{tabular}{@{}l|c|ccc|ccc|ccc@{}}
\toprule
Depth & - & - & - & - & 1e-3 & 5e-3 & 1e-2 & 1e-3 & 5e-3 & 1e-2 \\
Semantics  & - & 5\% & 10\% & 20\% & - & - & - & 5\% & 10\% & 20\% \\
\midrule
Sem. $\uparrow$  & 23.6 & 23.2 & 22.8 & 22.4 & 23.2 & 22.4 & 20.8 & 22.9 & 22.0 & 19.2 \\
Dyn $\uparrow$   & 21.7 & 20.1 & 19.7 & 19.0 & 20.8 & 20.3 & 19.4 & 20.4 & 19.5 & 16.3 \\
Occ. $\uparrow$ & 45.2 & 45.7 & 45.3 & 45.1 & 45.2 & 43.1 & 41.5 & 44.5 & 43.4 & 40.8 \\
\bottomrule
\end{tabular}
\label{tab:supmat:pseudo-label-noise}
\caption{RayIoU for noisy depth and/or semantic psuedo-labels.}
\end{table*}

\subsection{Supervision signals}
We compare two supervision strategies: direct 4D query-based supervision and an image-space rendering alternative that uses the same framework and supervision sources but applies losses via alpha compositing (\cref{sec:supmat:supervision}).
\Cref{tab:supmat:ablations:supervision} lists the full metrics. 
Rendering supervision achieves reasonable performance (slightly better than similar prior work \cite{huang2024selfocc, zhang2023occnerf, wang2024distillnerf}) but performs worse than query-based supervision across both semantic and geometric metrics. 
Adding feature distillation improves the query-based variant slightly, while combining rendering and query losses provides no additional benefit. 
This suggests that once explicit 4D supervision is available, image-space constraints become largely redundant.

\begin{table}[b]
\centering
\setlength{\tabcolsep}{2pt}
\renewcommand{\arraystretch}{0.9}
\begin{tabular}{
    *{4}{C{0.7cm}} 
    *{3}{C{0.7cm}} 
    *{3}{C{0.7cm}}
}
\toprule
\multicolumn{4}{c}{\textbf{Supervision Type}} &
\multicolumn{3}{c}{\textbf{RayIoU}$\uparrow$} &
\multicolumn{3}{c}{\textbf{IoU}$\uparrow$} \\ 
\cmidrule(lr){1-4}\cmidrule(lr){5-7}\cmidrule(lr){8-10}
\multicolumn{3}{c}{Query-based} &\multirow{2}{*}{Rend} & \multirow{2}{*}{\semantic} & \multirow{2}{*}{\dynamic} & \multirow{2}{*}{\occupancy} & \multirow{2}{*}{\semantic} & \multirow{2}{*}{\dynamic} & \multirow{2}{*}{\occupancy}  \\
Occ. & Sem. & Feat. & & & & & & &\\
\midrule
& & & \cmark & 15.0 & 9.8 & 41.7 & 13.8 & 9.8 & 53.8 \\
\cmark & \cmark & & & 23.6 & 21.7 & 45.2 & 21.3 & 13.2 & 55.0 \\
\cmark & \cmark & \cmark & & 23.9 & 22.3 & 45.0 & 21.5 & 13.5 & 55.5 \\ 
\cmark & \cmark & \cmark & \cmark & 23.3 & 21.3 & 45.6 & 21.2 & 13.1 & 55.1 \\
\bottomrule
\end{tabular}
\caption{Ablation of supervision signals. Image-space rendering supervision performs worse than 4D query-based supervision in both semantic and geometric accuracy within our framework.}
\label{tab:supmat:ablations:supervision}
\end{table}

\subsection{Supervision window}
Training supervision is derived from multiple adjacent time steps. 
\Cref{fig:temporal-window} shows that including nearby frames up to three seconds improves performance but degrades beyond that, suggesting that large temporal offsets introduce excessive difficulty for supervision.
Unlike GaussianFlowOcc, our model remains stable even for narrow windows, indicating it can extract reliable geometric cues from limited temporal context.
We apply supervision symmetrically in both forward and backward directions rather than predicting only the future.
This bidirectional setup effectively doubles the available training signals and, as shown in \cref{fig:temporal-window}, improves geometric reasoning around the reference frame.
Together, these experiments confirm that temporal supervision provides stronger geometric priors and improves semantic occupancy for the input frame.
\begin{figure}[!t]
    \centering
    \input{figures/temporal_window_fig.tex}
    \caption{Effect of temporal window. 
    Supervising across multiple timesteps improves geometric priors. 
    Forward and backward supervision perform better than just forward.}
    \label{fig:temporal-window}
\end{figure}

\subsection{Point cloud sampling strategies}
\Cref{tab:ablations:query_sampling} compares alternative strategies for selecting which points to use for pseudo and lidar supervision.
For lidar, moderate up weighting of dynamic points ($2$x–$5$x) improves dynamic-class performance by mitigating the inherent sparsity of moving objects. 
Higher weight to dynamic points beyond this offers no further benefit and can introduce instability.
Sampling points uniformly from a voxel-grid with cell length $0.4$m to match Occ3D slightly reduces performance.
In contrast, pseudo point clouds are already dense also for dynamic objects, making their performance largely insensitive to the sampling scheme; uniform sampling performs on par with dynamic up sampling.
For simplicity and reproducibility, we therefore adopt uniform downsampling.

\begin{table}[t]
\centering
\setlength{\tabcolsep}{2pt}
\renewcommand{\arraystretch}{0.9}
\begin{tabular}{
    L{1.5cm} 
    C{0.8cm} C{0.8cm} C{0.8cm} 
    C{0.8cm} C{0.8cm} C{0.8cm}
}
\toprule
\multirow{2}{*}{\textbf{Sampling}} &
\multicolumn{3}{c}{\textbf{Pseudo}} &
\multicolumn{3}{c}{\textbf{Lidar}} \\
& \semantic & \dynamic & \occupancy & \semantic & \dynamic & \occupancy \\
\midrule
Uniform     & 23.6 & 21.7 & 45.2 & 22.9 & 19.4 & 48.3 \\
Dyn.~2×     & 23.9 & 22.1 & 45.1 & 23.4 & 20.6 & 48.1 \\
Dyn.~5×     & 23.3 & 21.2 & 44.5 & 24.3 & 22.1 & 47.8 \\
Dyn.~10×    & 23.6 & 22.0 & 45.3 & 23.8 & 21.6 & 47.0 \\
Voxel-uni   & 22.6 & 20.2 & 44.5 & 21.6 & 18.8 & 47.7 \\
\bottomrule
\end{tabular}
\caption{RayIoU$\uparrow$ for query sampling ablation under pseudo and \lidar supervision. It is more important to upsample dynamic points when using \lidar than pseudo-point clouds.}
\vspace{12pt}
\label{tab:ablations:query_sampling}
\end{table}

\subsection{Cell size variations}
\begin{figure}[t]
  \centering
  \input{figures/cell_size_scaling.tex}
  \vspace{-16pt}
  \caption{Effect of BEV cell size on RayIoU$\uparrow$ metrics. Smaller cells leads to a notable improvements in performance, underscoring the need for an efficient, compact, and fine-grained BEV representation.
  In our framework the performance plateaus when using cell sizes smaller than $0.16$.}
  \vspace{3.5mm}
  \label{fig:cell_size_plot}
\end{figure}
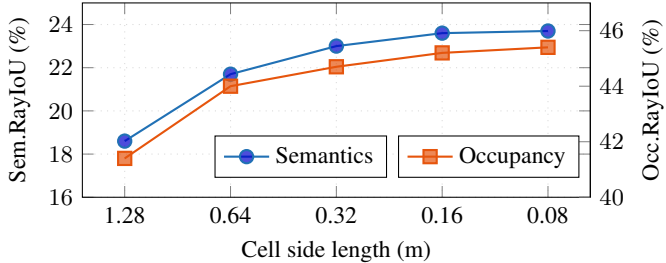

To isolate the effect of BEV resolution, \cref{fig:cell_size_plot} evaluates semantic and occupancy RayIoU across cell sizes. 
Performance improves as resolution increases, showing that coarse BEV grids can be a bottleneck.
We hypothesis that large cells may smear thin structures, inflate surfaces, and degrade the decoder’s ability to localize geometry from queries. 
However, beyond roughly $0.16$m, the improvements plateau and increasing resolution further yields negligible performance changes while growing memory and latency. 
This plateau justifies our design choice using $0.16$m resolution near the ego vehicle, but contraction is required to maintain this resolution without exploding memory using long range.

\subsection{Compression range sensitivity}
\begin{table}[h]
\centering
\setlength{\tabcolsep}{3.5pt}
\renewcommand{\arraystretch}{0.7}
\vspace{-1.5mm}
\begin{tabular}{@{}l|cccc@{}}
\toprule
$d_{\text{near}}$, $K_{\text{hr}}$   & 40m & 30m & 20m & 10m \\ \midrule 
Sem. $\uparrow$ & 23.6 & 23.7 & 23.5 & 23.4 \\ 
Dyn. $\uparrow$ & 21.7 & 21.4 & 21.2 & 21.1 \\ 
Occ. $\uparrow$ & 45.2 & 45.7 & 45.8 & 45.4 \\ 
\bottomrule
\end{tabular}
\caption{RayIoU exhibits low sensitivity to parameter choice}
\label{tab:supmatcompressionrange}
\vspace{3.5mm}
\end{table}
\cref{tab:supmatcompressionrange} shows low sensitivity to $d_{\text{near}}$ and $K_{\text{hr}}$, indicating that contracted BEV features preserve sufficient information to maintain performance.

\subsection{Increasing dataset size}
\label{sec:supmat:datasetsize}
We further study scaling by adding Argoverse 2 and the NAVSIM split of nuPlan~\cite{Argoverse2, dauner2024navsim, nuplan} to the training data mix.
To reduce preprocessing costs, we assume semantic pseudo-labels are available only for nuScenes, while the additional datasets provide \lidar occupancy and VFM feature supervision.
As shown in \Cref{tab:ablations:datasets}, adding either AV2 or nuPlan improves mean RayIoU by approximately +2 points over the nuScenes-only baseline, demonstrating that \modelname can leverage additional data even under weaker and heterogeneous supervision. Notably, nuPlan yields larger gains than AV2, likely due to its scale and closer distribution to nuScenes.
However, combining all three datasets does not yield further improvements, indicating diminishing returns from additional data. 
Together with \Cref{tab:datasets-pseudo}, this suggests that performance is ultimately limited by supervision quality: while additional data provides initial gains, weaker VFM-based supervision constrains further improvement compared to pseudo-label supervision.

\begin{table}
    \centering
    \setlength{\tabcolsep}{3pt} 
    \renewcommand{\arraystretch}{0.7} 
    \begin{tabular}{
        C{0.7cm} C{0.7cm} C{0.7cm}  
        C{0.9cm} C{0.9cm} C{0.9cm}  
    }
    \toprule
    \multicolumn{3}{c}{\textbf{Datasets}} & \multicolumn{3}{c}{\textbf{RayIoU}$\uparrow$} \\
    \cmidrule(lr){4-6} \cmidrule(lr){1-3}
    nuSc & AV2 & nuPl & \semantic & \dynamic & \occupancy \\
    \midrule
    \cmark &        &           & 22.9 & 19.4 & 48.3 \\
    \cmark & \cmark &           & 24.3 & 20.8 & 50.5 \\
    \cmark &        & \cmark    & 25.3 & 22.3 & 50.3 \\
    \cmark & \cmark & \cmark    & 25.0 & 22.3 & 50.1 \\
    \bottomrule
    \end{tabular}
    \caption{Effect of combining additional datasets with nuScenes.}
    \label{tab:ablations:datasets}
\end{table}

\subsection{\Lidar segmentation}
\begin{figure*}[h]
    \centering
    \begin{subfigure}[t]{0.48\linewidth}
        \centering
        \includegraphics[width=\linewidth]{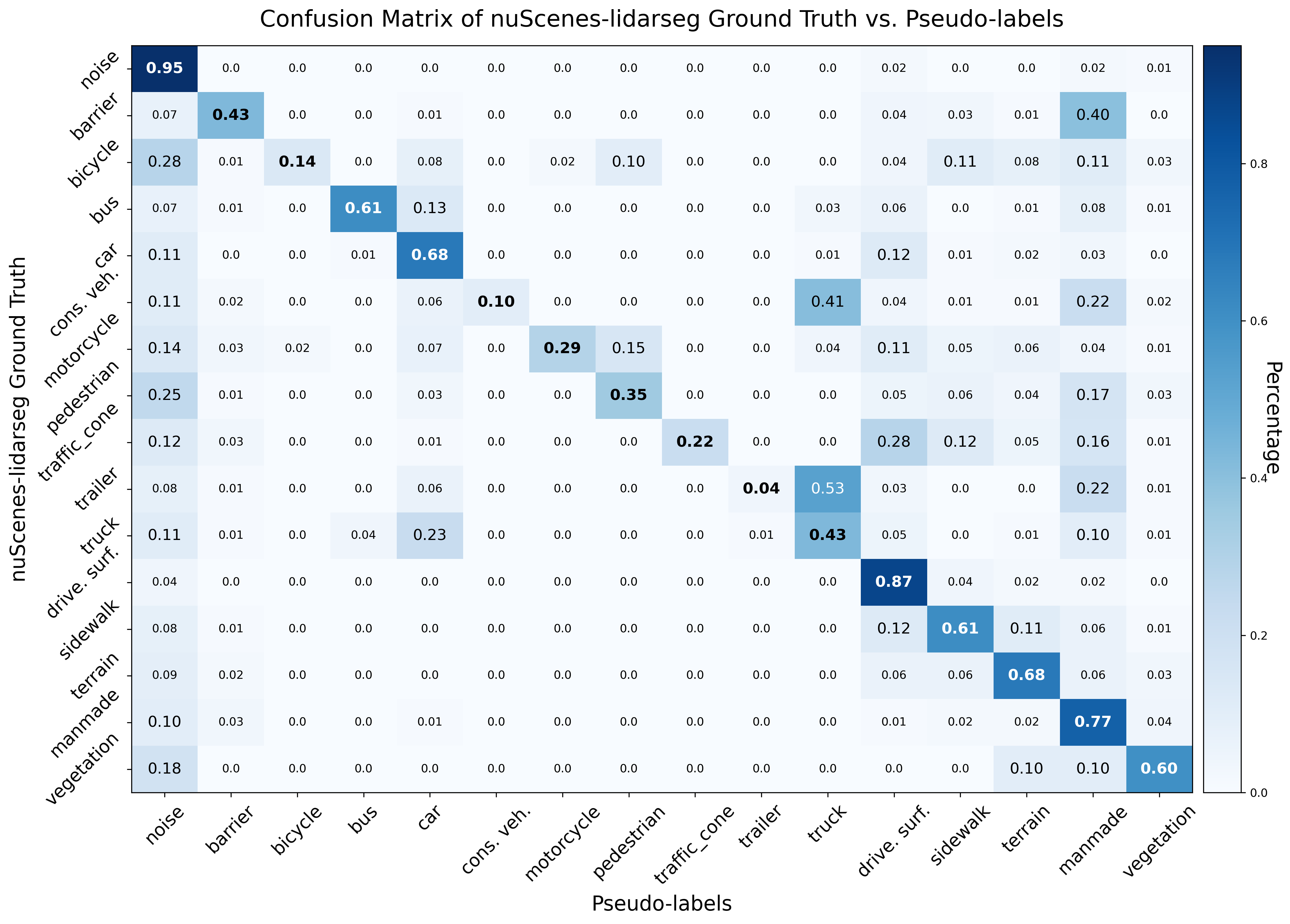}
        \caption{Pseudo-labels}
        \label{fig:sub1}
    \end{subfigure}
    \hfill
    \begin{subfigure}[t]{0.48\linewidth}
        \centering
        \includegraphics[width=\linewidth]{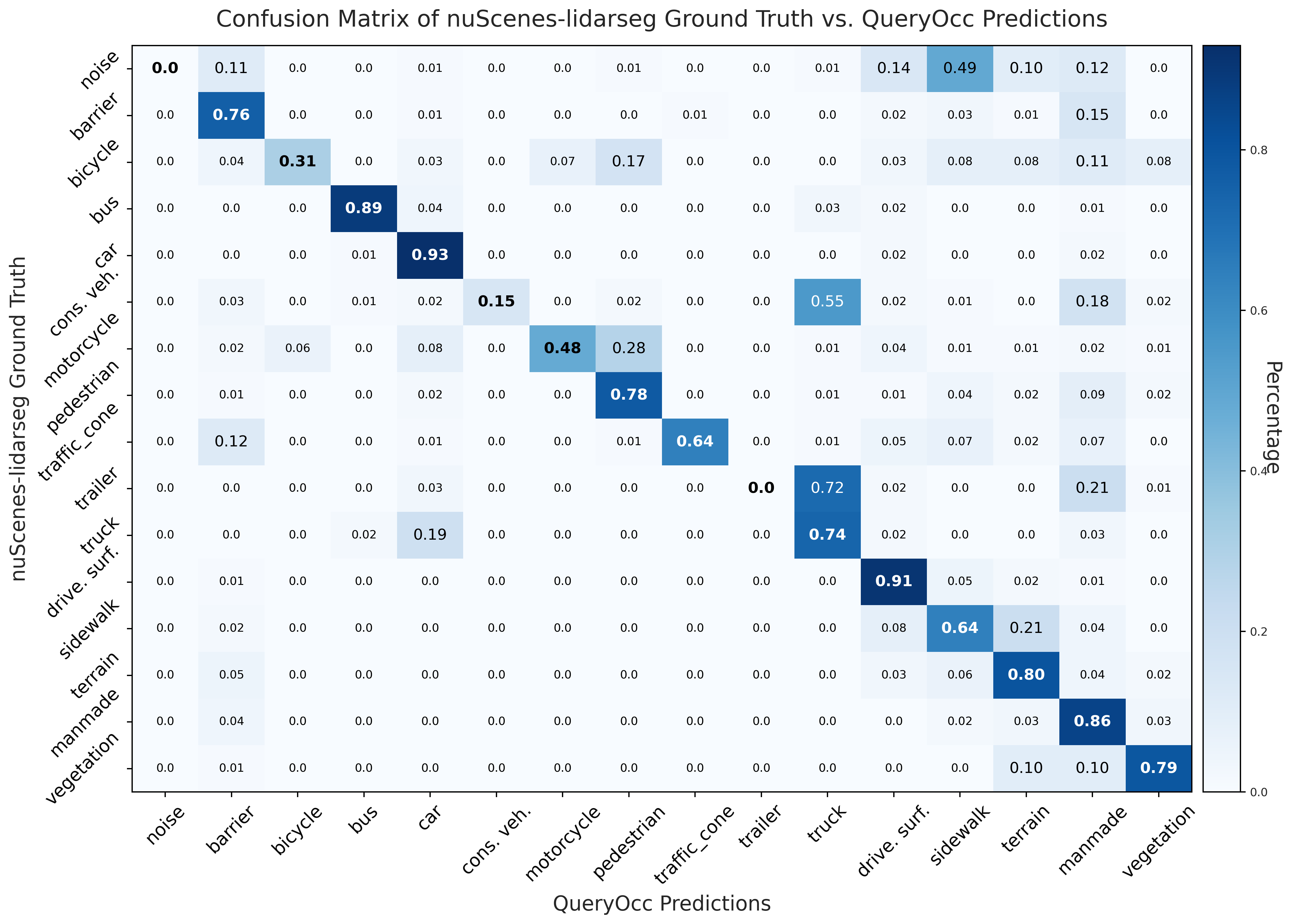}
        \caption{\modelname}
        \label{fig:sub2}
    \end{subfigure}
    \vspace{12pt}
    \caption{Confusion matrices for the relevant classes on the nuScenes lidar segmentation dataset \cite{fong2022panoptic}.}
    \vspace{8mm}
    \label{fig:supmat:lidarsegconfusionmatrix}
\end{figure*}

We leverage the \lidar segmentation annotations introduced in Panoptic nuScenes~\cite{fong2022panoptic} to analyze the semantic quality of both the pseudo-labels, and the predictions of \modelname trained using lidar point cloud supervision. 
These annotations provide per-point ground truth labels, enabling fine-grained evaluation beyond voxel-based metrics.
For evaluation, we query \modelname at the 3D locations of the \lidar points and compare the predicted semantic classes to the ground truth labels. 
We apply the same projection-based procedure used to construct pseudo-labels to obtain comparable predictions from the pseudo-labels for the full point cloud.
\cref{tab:supmat:lidarsegmetrics} shows that \modelname consistently outperforms the pseudo-labels across all reported metrics. 
A more detailed view in \cref{tab:supmat:lidarseg-iou} reveals that per-class intersection-over-union improves for all categories except \emph{trailer}, indicating that the model does not simply replicate the supervision signal but systematically enhances it.
This trend is further supported by the confusion matrices in \Cref{fig:supmat:lidarsegconfusionmatrix}, where \modelname exhibits reduced inter-class confusion compared to the pseudo-labels, particularly among major semantic groups. 
However, ambiguities between semantically similar classes, such as \emph{truck} and \emph{trailer}, persist for both.
This indicates that \modelname is able to denoise and refine the supervision signal, despite being trained on imperfect pseudo-labels.

\begin{table}[h]
\centering
\setlength{\tabcolsep}{3pt}
\renewcommand{\arraystretch}{0.7}

\vspace{-1mm}
\begin{tabular}{
    L{2.2cm} |
    *{4}{C{1.0cm}} 
}
\toprule
Method & IoU & Rec. & F1 & Acc. \\
\midrule
Pseudo-labels & 35.5 & 45.0 & 48.6 & 45.0  \\
\modelname   & 49.8 & 64.5 & 62.0 & 64.5  \\
\bottomrule
\end{tabular}
\caption{QueryOcc performs better on lidar segmentation metrics than the pseduo-ground truth.}
\label{tab:supmat:lidarsegmetrics}
\end{table}

\begin{table*}
\centering
\setlength{\tabcolsep}{2pt}
\renewcommand{\arraystretch}{1.05}

\newcommand{\clsname}[2]{%
  \rotatebox{70}{\textcolor{#2}{$\blacksquare$}~#1}%
}

\begin{tabular}{
  l
  *{16}{c}
}
\toprule
\textbf{Source} &
\textbf{Mean} &
\clsname{barrier}{barrier} &
\clsname{bicycle}{bicycle} &
\clsname{bus}{bus} &
\clsname{car}{car} &
\clsname{cons. veh.}{construct} &
\clsname{motorcycle}{motor} &
\clsname{pedestrian}{pedestrian} &
\clsname{traffic cone}{traffic} &
\clsname{trailer}{trailer} &
\clsname{truck}{truck} &
\clsname{drive. surf.}{driveable} &
\clsname{sidewalk}{sidewalk} &
\clsname{terrain}{terrain} &
\clsname{manmade}{manmade} &
\clsname{vegetation}{vegetation} \\
\midrule

Pseudo-labels &
35.5 & 19.1 & 9.5 & 48.5 & 55.6 & 9.2 & 23.2 & 28.5 & 15.6 & 4.1 & 32.8 & 79.3 & 43.6 & 46.3 & 62.5 & 54.2 \\

QueryOcc &
49.8 & 27.6 & 16.4 & 78.3 & 80.7 & 15.1 & 46.0 & 55.9 & 32.3 & 0.3 & 62.0 & 87.9 & 44.0 & 53.6 & 74.5 & 73.2 \\

\bottomrule
\end{tabular}
\caption{Per-class IoU (\%) against lidar segmentation ground truth.}
\vspace{12pt}
\label{tab:supmat:lidarseg-iou}
\end{table*}

\section{Qualitative examples}
\label{sec:supmat:qualitative:examples}
\begin{figure*}
    \centering
    \includegraphics[width=0.8\linewidth]{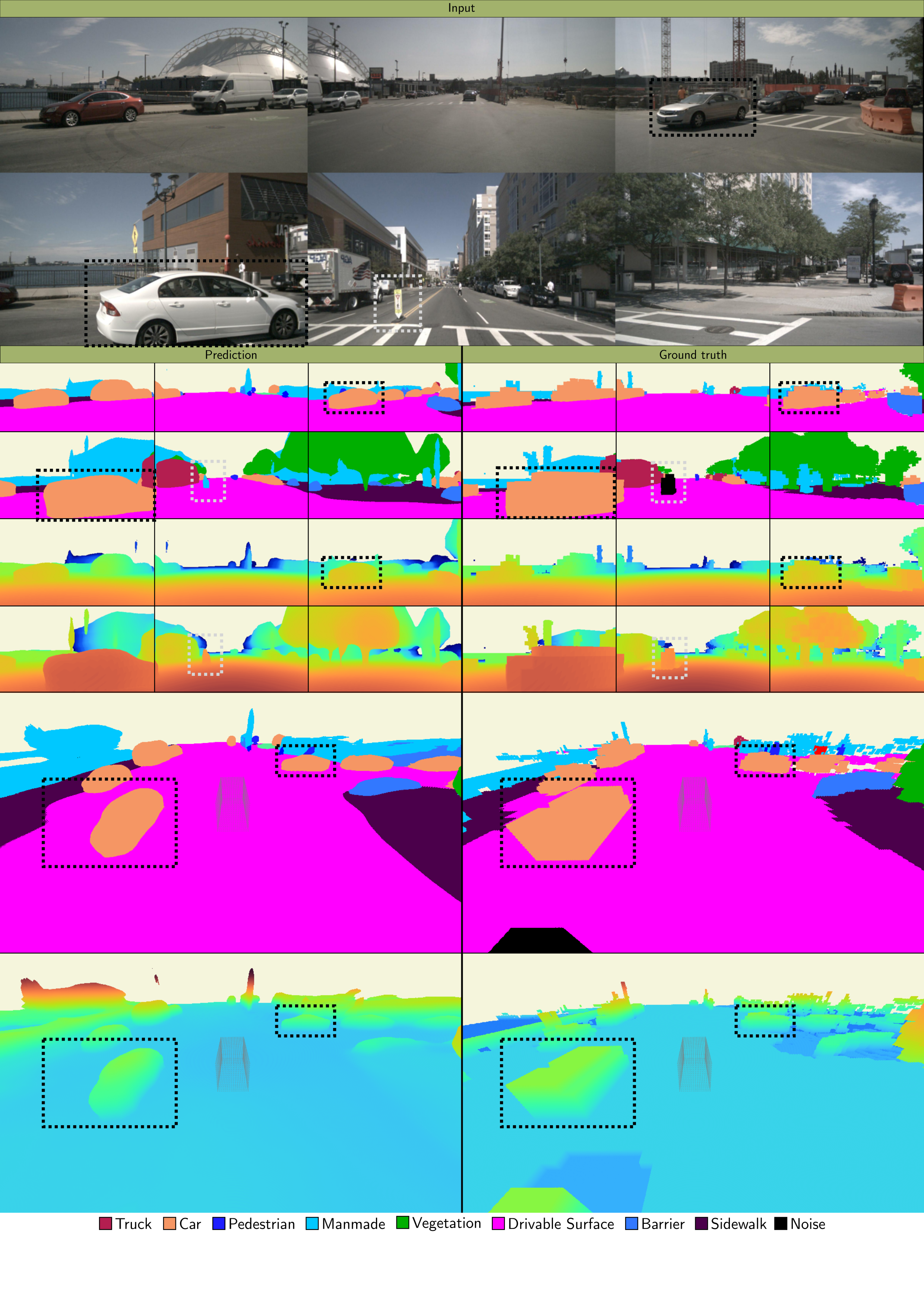}
    \vspace{-30pt}
    \caption{\textbf{Scene A:} Qualitative example from \modelname. 
    The model reconstructs vehicle geometry with high fidelity (black boxes) and recovers small narrow structures such as road signs (white box). 
    Semantic and occupancy predictions remain spatially consistent across camera views and exhibit smoother geometry than the coarse voxel grid from the Occ3D ground truth.}
    \label{fig:qual-a}
\end{figure*}

\begin{figure*}
    \centering
    \includegraphics[width=0.8\linewidth]{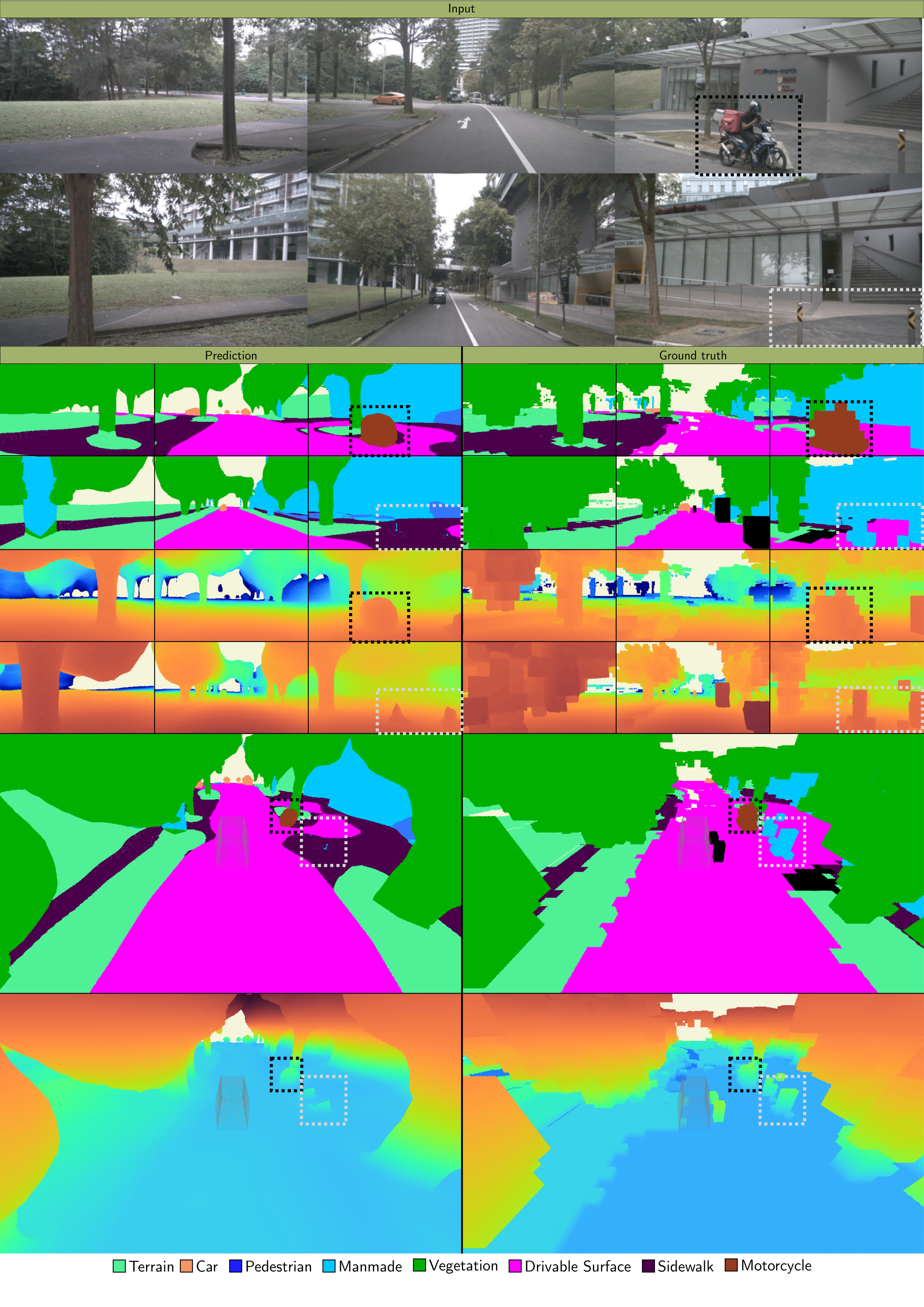}
    \vspace{-12pt}
    \caption{\textbf{Scene B:} The motorcyclist is correctly identified (black box), while thin manmade poles are missed semantically despite partial occupancy signals (white box), illustrating the difficulty of extremely narrow structures. 
    Broader scene elements such as sidewalk, terrain, and vegetation are predicted coherently.}
    \label{fig:qual-b}
\end{figure*}

\begin{figure*}
    \centering
    \includegraphics[width=0.8\linewidth]{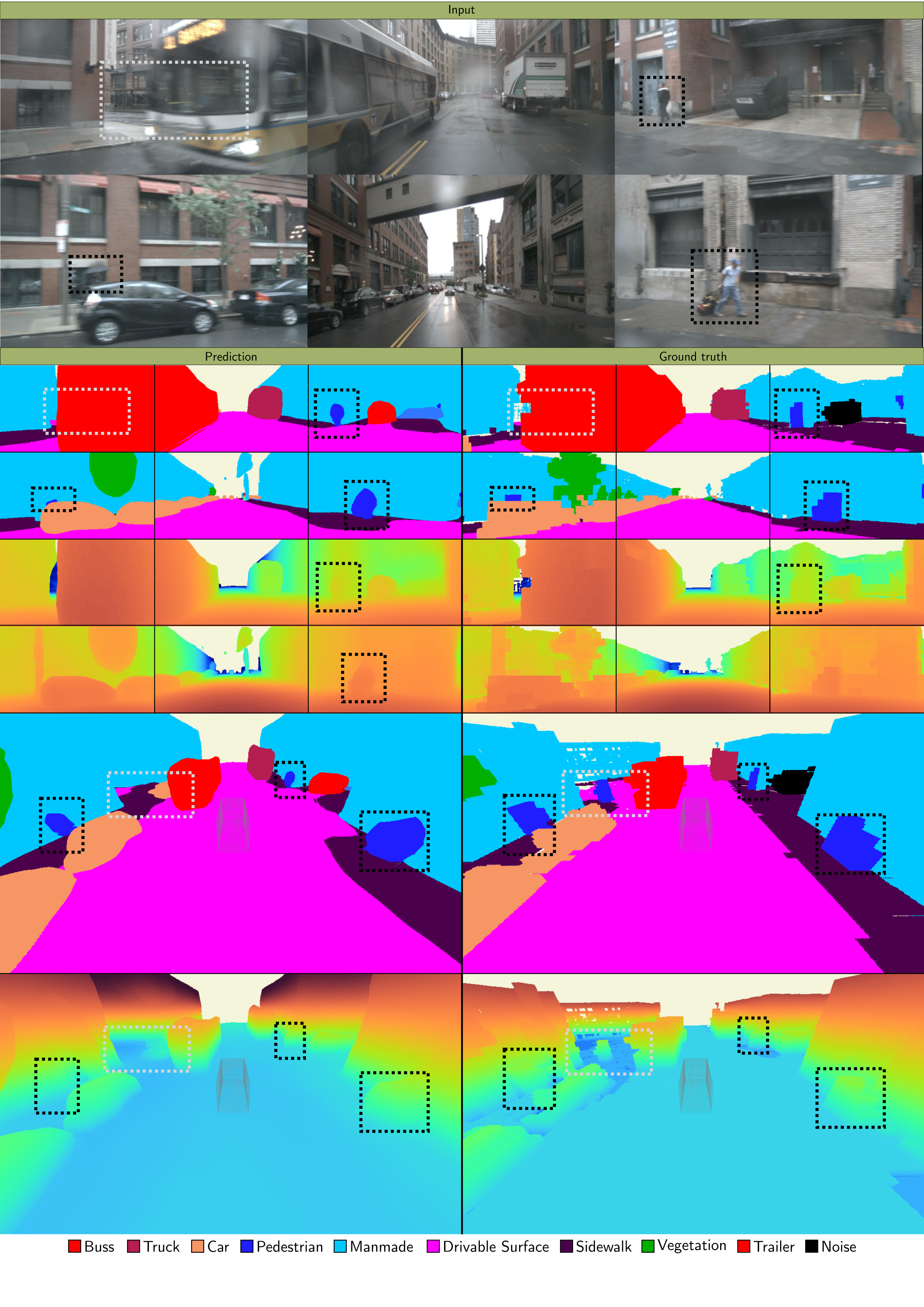}
    \vspace{-15pt}
    \caption{\textbf{Scene C:} Pedestrians (black boxes) are reconstructed with high consistency, including partially occluded instances. 
    A fully occluded pedestrian behind a bus is missed (white box), though the model infers plausible surface continuation behind an occlusion. 
    The predicted occupancy also reflects common priors such as possible parked vehicles in similar street configurations.}
    \label{fig:qual-c}
\end{figure*}

We present qualitative examples from \modelname in \cref{fig:qual-a,fig:qual-b,fig:qual-c} and separate videos.
Overall, these examples show that \modelname produces sharp geometry, maintains fine-grained detail, and infers plausible structures behind occlusions. 

\parsection{Scene A (\cref{fig:qual-a})}
Vehicle surfaces appear sharper and less boxy than the voxelized ground-truth occupancy, which is visibly inflated due to the $0.4$ m discretization (black boxes).
Continuous-field self-supervised methods (including \modelname and \eg GaussianFlowOcc), are typically not constrained by this voxel resolution and naturally produce higher-frequency details than the grid can represent.
\modelname also recovers small, narrow structures such as road signs (white box), showing that the contracted BEV retains sufficient local detail for fine-scale prediction.

\parsection{Scene B (\cref{fig:qual-b})}
The motorcyclist on the right is correctly detected (black box), while nearby thin poles (white box) are missed. 
This reflects class imbalance and the difficulty of recovering extremely narrow structures. 
We note that the occupancy yields some structure for the bottom of the poles, but due to the class-uncertainty no semantic prediction is produced.
Larger background regions such as terrain, sidewalk, vegetation are estimated well.

\parsection{Scene C (\cref{fig:qual-c})}
Pedestrians (black boxes) are accurately reconstructed, including the partially occluded walker with an umbrella in the left rear camera.
The method naturally misses a pedestrian fully occluded by a bus (white box), but still infers plausible surface structure, \ie drivable surface and sidewalk, behind the occlusion. 
The predicted occupancy also suggests a possible car, which likely reflects dataset bias toward parked vehicles in similar configurations.
We hypothesize that temporal supervision across adjacent frames, in which the occluding objects move and expose new areas, encourages the model to infer plausible structures behind occlusions.

\begin{figure*}
    \centering
    \includegraphics[width=0.8\linewidth]{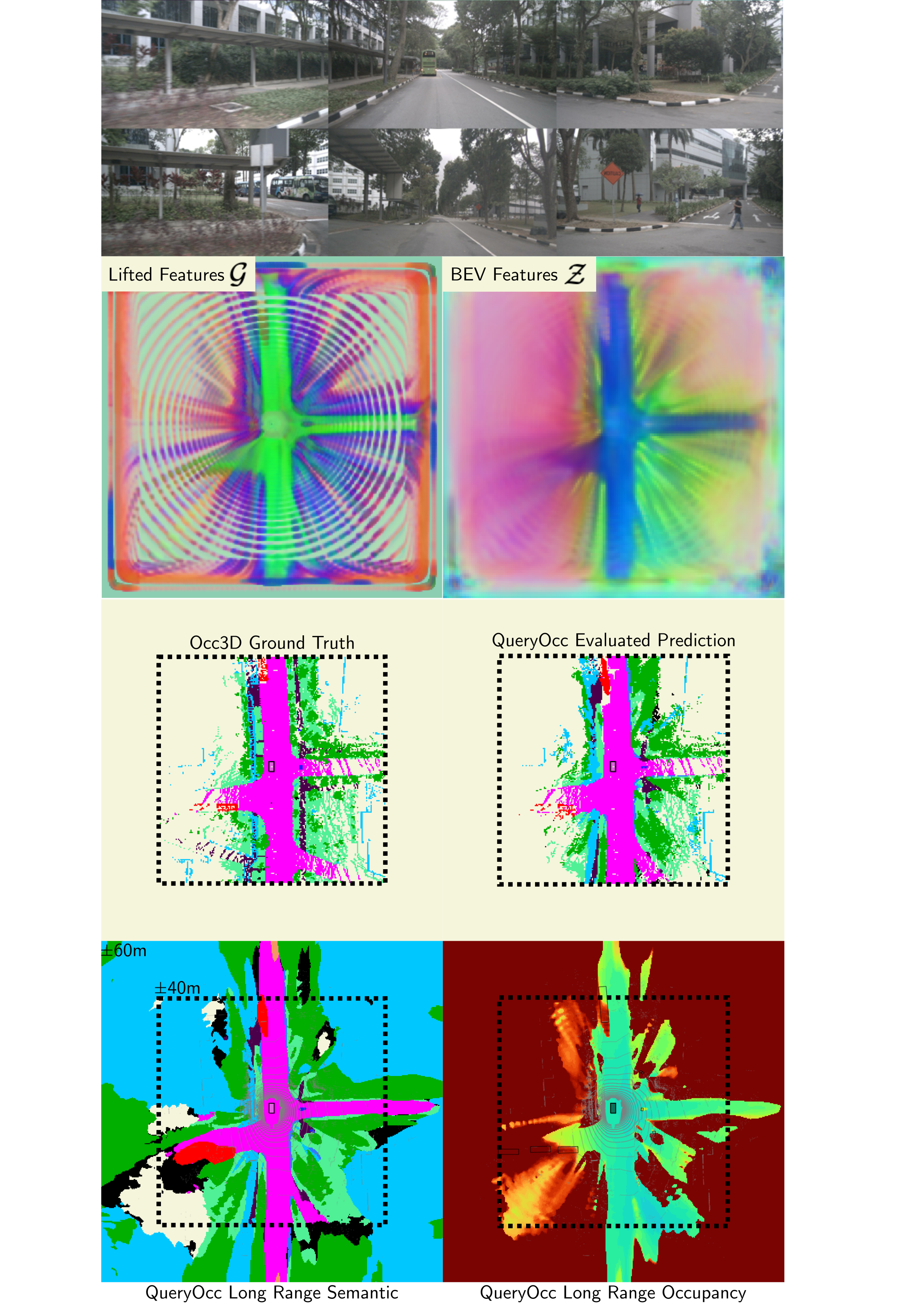}
    \caption{\textbf{Scene D:}
    Predictions beyond the Occ3D evaluation range.
    We visualize lifted features, BEV features, predictions, and ground truth, extending the view from the standard $\pm 40$ m to $\pm 60$m.
    The decoder reconstructs plausible geometry both inside the high-resolution region and in the contracted far-field area (black dashed box).
    Road layout and free-space structure remain consistent even well outside the evaluation boundary.
    }
    \label{fig:long-range-b}
\end{figure*}

\parsection{Long-range Scene D (\cref{fig:long-range-b})}
We extend the visualization beyond the $\pm 40$ m Occ3D evaluation range to $\pm 60$ m to assess how the model behaves in the contracted far-field region (outside the black dashed box).
Notably, the contracted region still provides sufficient spatial signal for the decoder to reconstruct consistent surfaces, showing that the representation retains usable geometric information even after contraction.

\begin{figure*}
    \centering
    \includegraphics[width=0.8\linewidth]{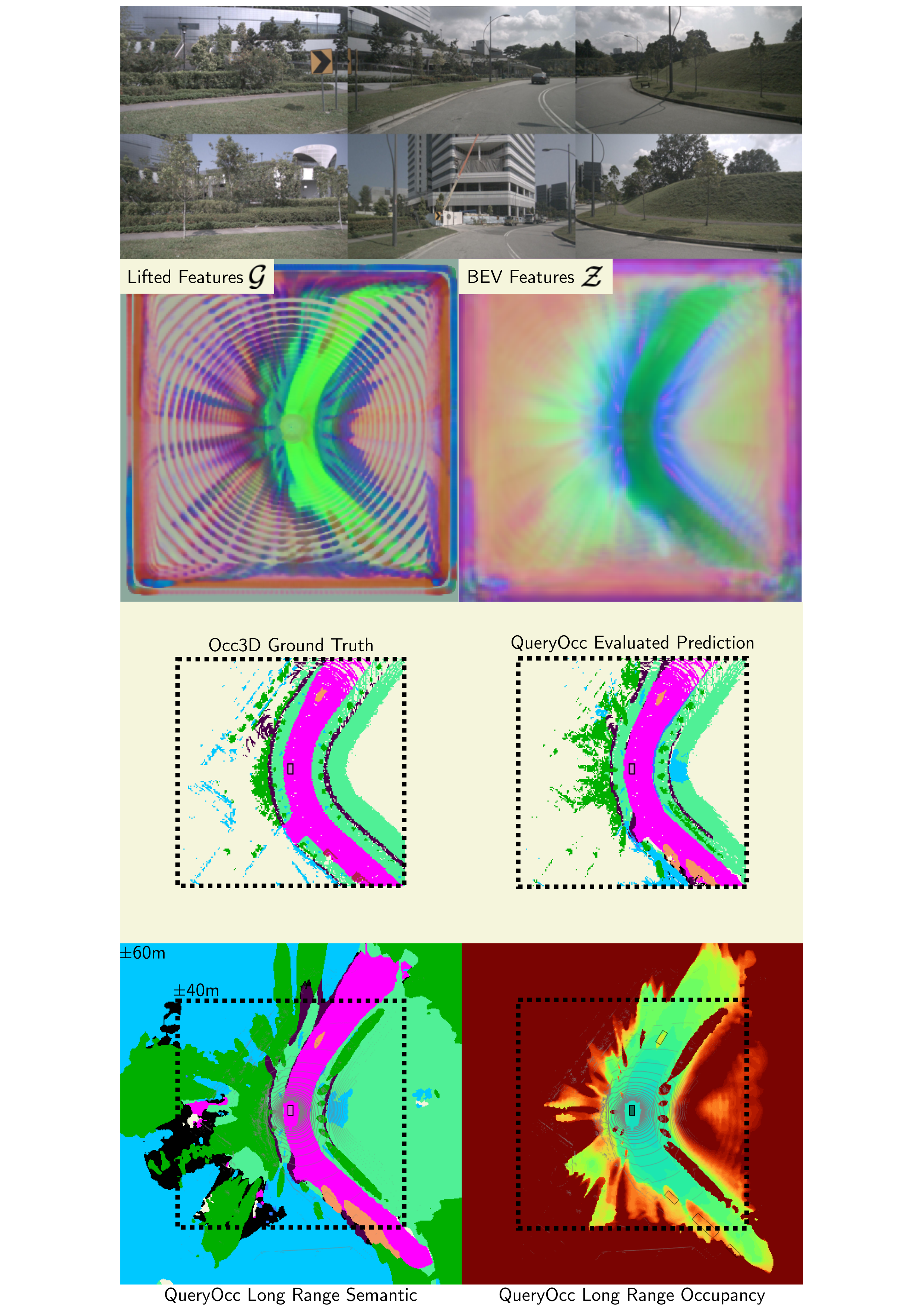}
    \caption{
    \textbf{Scene E:}
    Geometric consistency in the contracted far-field.
    We again extend visualization to $60$m range.
    The decoder recovers the bending road and surrounding free-space structure from the contracted region outside the high-resolution area (black dashed box), demonstrating that the representation preserves spatial cues even at long distances.
    }
    \label{fig:long-range-c}
\end{figure*}

\parsection{Long-range Scene E (\cref{fig:long-range-c})}
This example illustrates the model’s ability to recover structure in a more complex far-field scenario.
Although the road bends outside the high-resolution area, the decoder still reconstructs its curvature from contracted BEV features, indicating that the contraction preserves directional and structural cues at long ranges.
The predictions remain geometrically stable up to $60$m, suggesting that the model generalizes beyond the evaluation boundary and can leverage contracted features effectively.

\section{Continuous Predictions vs.\ Voxelized Ground Truth}
\label{sec:supmat:discussion}
The qualitative examples reveal a systematic discrepancy between our predictions and the voxelized ground truth used in Occ3D.
The Occ3D ground truth is constructed by fusing multiple lidar sweeps into a fixed $0.4$,m grid, which inflates geometry, thickens object boundaries, and removes fine-grained structure.

Continuous-field methods, including ours and rendering-based approaches such as GaussianFlowOcc, learn a smooth occupancy function in $\mathbb{R}^3$ and are not restricted by voxel resolution.
They can therefore produce sharper transitions and thin structures that the discretized labels cannot represent.
These representational differences can make predictions appear visually higher-resolution than the voxel grid, particularly around vehicles, poles, trees, and other narrow structures, without leading to increased metric scores.
This mismatch also motivates the use of RayIoU, which is less sensitive to voxel inflation and better aligned with the underlying geometry.

Overall, the discrepancy is inherent to voxel-based supervision and evaluation.
Continuous 4D methods generate finer spatial detail than the evaluation grid can express, which can put them at a disadvantage under voxel-based metrics even when the underlying geometry is consistent.

%% file: figures/temporal_window_fig.tex
\begin{tikzpicture}
\pgfplotstableread{
Window   GaussianFlowOcc   Forward   ForwardBackward   LidarForwardBackward
0.0      12.67             18.8      18.8              18.4
0.5      13.42             19.0      20.5              19.4
1.0      14.17             19.3      21.1              19.6
2.0      14.71             19.5      20.9              19.4
3.0      17.08             20.1      21.0              19.8
4.0      14.82             19.8      20.2              20.0
}\temporalData

\begin{axis}[
  width=\columnwidth, height=0.4\columnwidth,
  xlabel={Temporal Window (s)},
  ylabel={mIoU},
  ylabel style={yshift=-6pt},
  xmin=0, xmax=4,
  ymin=10, ymax=22,
  xtick={0,0.5,1,2,3,4},
  grid=both,
  line width=0.8pt,
  mark size=1.8pt,
  every axis/.append style={
  tick label style={font=\footnotesize},
  label style={font=\footnotesize},
  title style={font=\footnotesize}
},
  legend style={
    font=\footnotesize,
    at={(0.5,1.05)},
    anchor=south,
    legend columns=-1,
    draw=none            
  },
  legend cell align=left,
  every axis plot/.append style={thick},
]

\addplot [
  gray!70,
  solid,
  mark=*,
  line width=0.8pt,
  mark size=1.8pt,
  mark options={fill=gray!70, line width=0.3pt}
]
table [x=Window, y=GaussianFlowOcc] {\temporalData};
\addlegendentry{GaussianFlowOcc};

\addplot [
  color={rgb,255:red,33; green,113; blue,181},
  dashed,
  mark=square*,
  line width=0.8pt,
  mark size=1.8pt,
  mark options={fill={rgb,255:red,33; green,113; blue,181}, line width=0.3pt}
]
table [x=Window, y=Forward] {\temporalData};
\addlegendentry{Forward};

\addplot [
  color={rgb,255:red,33; green,113; blue,181},
  solid,
  mark=*,
  line width=0.8pt,
  mark size=1.8pt,
  mark options={fill={rgb,255:red,33; green,113; blue,181}, line width=0.3pt}
]
table [x=Window, y=ForwardBackward] {\temporalData};
\addlegendentry{Forward-Backward};


\end{axis}
\end{tikzpicture}

%% file: figures/cell_size_scaling.tex
\pgfplotstableread{
Voxel mean binary
0.08    23.7    45.4
0.16	23.6	45.2
0.32	23.0	44.7
0.64	21.7	44.0
1.28	18.6	41.4
}\datatable

\begin{tikzpicture}
  \begin{axis}[
    width=\columnwidth,
    height=\columnwidth/2,
    xlabel={Cell side length (m)},
    ylabel={\semantic RayIoU (\%)},
    x dir=reverse,
    xtick=data,
    xmode=log,
    log basis x=2,
    xticklabels from table={\datatable}{Voxel},
    ymin=16, ymax=25,
    grid=both,
    grid style={dotted,gray!30},
    tick label style={font=\small},
    label style={font=\small},
    legend style={at={(0.43,0.1)},anchor=south,legend columns=-1,font=\small},
    every axis plot/.append style={thick,mark size=2.5pt},
  ]
    \addplot+[mark=*,
      color={rgb,255:red,33; green,113; blue,181},
      fill opacity=0.7]
      table[x=Voxel, y=mean] {\datatable};
    \addlegendentry{Semantics}
  \end{axis}

  \begin{axis}[
    width=\columnwidth,
    height=\columnwidth/2,
    x dir=reverse,
    axis y line*=right,
    axis x line=none,
    xmode=log,
    log basis x=2,
    ylabel={\occupancy RayIoU (\%)},
    ymin=40, ymax=47,
    tick label style={font=\small},
    label style={font=\small},
    legend style={at={(0.8,0.1)},anchor=south,legend columns=-1,font=\small},
    every axis plot/.append style={thick,mark size=2.5pt},
  ]
    \addplot+[mark=square*,
      color={rgb,255:red,230; green,85; blue,13},
      mark options={fill={rgb,255:red,230; green,85; blue,13}},
      fill opacity=0.7]
      table[x=Voxel, y=binary] {\datatable};
    \addlegendentry{Occupancy}
  \end{axis}
\end{tikzpicture}